\pdfoutput=1

\documentclass[11pt]{article}

\usepackage[final]{acl}

\usepackage{times}
\usepackage{graphicx}
\usepackage{latexsym}
\usepackage{algorithm}
\usepackage{multirow}
\usepackage{algpseudocode}

\usepackage[T1]{fontenc}

\usepackage[utf8]{inputenc}

\usepackage{microtype}

\usepackage{inconsolata}

\usepackage{graphicx}
\usepackage{amsmath}
\usepackage{amssymb}
\usepackage[most]{tcolorbox}
\tcbuselibrary{listings, breakable}
\definecolor{lightgray}{gray}{0.97}

\newtcolorbox{promptbox}{
  enhanced,
  colback=lightgray,
  colframe=black,
  fontupper=\ttfamily\footnotesize,
  boxrule=0.4pt,
  arc=2mm,
  outer arc=2mm,
  breakable,
  left=2mm,
  right=2mm,
  top=1mm,
  bottom=1mm,
}

\title{\textsc{MuseScorer}: Idea Originality Scoring At Scale}

\author{Ali Sarosh Bangash$^{*}$, Krish Veera$^{*}$, Ishfat Abrar Islam, Raiyan Abdul Baten$^{\dagger}$ \\
        Bellini College of Artificial Intelligence, Cybersecurity, and Computing,\\ University of South Florida, USA\\
        \texttt{\{alibangash, krishv, ishfatabrar, rbaten\}@usf.edu}\\
        \small{
       $^{*}$Equal contributions \quad
   $^{\dagger}$Correspondence: \href{mailto:rbaten@usf.edu}{rbaten@usf.edu}
 }}

\begin{document}
\maketitle
\begin{abstract}
An objective, face-valid method for scoring idea originality is to measure each idea’s statistical infrequency within a population---an approach long used in creativity research. Yet, computing these frequencies requires manually bucketing idea rephrasings, a process that is subjective, labor-intensive, error-prone, and brittle at scale. We introduce \textsc{MuseScorer}, a fully automated, psychometrically validated system for frequency-based originality scoring. \textsc{MuseScorer} integrates a Large Language Model (LLM) with externally orchestrated retrieval: given a new idea, it retrieves semantically similar prior idea-buckets and zero-shot prompts the LLM to judge whether the idea fits an existing bucket or forms a new one. These buckets enable frequency-based originality scoring without human annotation. Across five datasets ($N_{\text{participants}}{=}1143$, $n_{\text{ideas}}{=}16{,}294$), \textsc{MuseScorer} matches human annotators in idea clustering structure (AMI $=0.59$) and participant-level scoring ($r = 0.89$), while demonstrating strong convergent and external validity. The system enables scalable, intent-sensitive, and human-aligned originality assessment for creativity research.
\end{abstract}

\section{Introduction}

Assessing creativity at scale remains a central challenge in cognitive science and computational linguistics. Creativity scoring broadly considers two complementary dimensions: the \textit{intrinsic} qualities of ideas (e.g., creative ideas are semantically flexible or diverse) and their \textit{extrinsic} statistical infrequency within a population (i.e., original ideas do not appear very often)~\citep{beketayev2016scoring,runco2012standard}. Recent methods have scaled intrinsic assessments using unsupervised, semi-supervised, and supervised approaches~\citep{organisciak2023beyond,beaty2021automating,organisciak2020open}. In contrast, frequency-based originality scoring still depends on manual tabulation of response occurrences~\citep{reiter2019scoring}. This requires grouping rephrasings of the same idea into buckets (e.g., `hold papers down' and `use as a paperweight' for a brick), a process that is subjective, fatigue-intensive, and error-prone as annotators track an ever-expanding set of buckets~\citep{acar2014assessing,baten2020creativity,baten2021cues,baten2022novel,buczak2023machines}. Moreover, the field lacks consensus on what counts as an `infrequent' idea, leaving frequency-based scoring with limited psychometric validation.

We introduce \textsc{MuseScorer}, a fully automated, psychometrically validated system for frequency-based originality scoring---bringing us closer to comprehensive automated creativity assessment. Bucketing different phrasings of the same idea together is computationally non-trivial: (i)~semantic similarity alone is insufficient to distinguish rephrasings from distinct intents, (ii)~traditional clustering algorithms struggle with singleton and low-frequency ideas that are critical for infrequency scoring, (iii)~real-world idea datasets follow fat-tailed bucket size distributions, defying uniform or Gaussian assumptions, and (iv)~bucket count grows as new ideas arrive, rendering ineffective text labeling tools that require label sets apriori. \textsc{MuseScorer} addresses these challenges via an LLM-as-a-judge framework with externally orchestrated retrieval: for each new idea, it retrieves from its database semantically similar prior buckets as candidates and zero-shot prompts the LLM to decide whether the idea fits an existing bucket or forms a new one. Unlike conventional clustering, this approach replicates the granularity of human bucketing in both structure and resolution.

Our work also contributes to the creativity literature in two ways. \textit{First}, we establish rigorous psychometric validity for frequency-based originality scoring, showing high agreement with human annotations and strong correlations with relevant cognitive traits. In doing so, we elucidate how `infrequency' can be reliably operationalized. \textit{Second}, we release an automated, interpretable scoring pipeline deployable across diverse open-ended ideation tasks, enabling creativity research at scale\footnote{\texttt{https://github.com/cssai-research/MuseScorer}}. More broadly, \textsc{MuseScorer} demonstrates how advanced NLP methods can address long-standing annotation challenges, providing validated tools that adjacent disciplines can adopt with confidence.

\section{Related Work}

\subsection{Computational Assessment of Creativity}

Creativity assessment has long relied on divergent thinking tasks like the Alternate Uses Test (AUT), where participants list novel uses for everyday objects~\citep{guilford1967nature}. Response sets are then scored for fluency (idea count), flexibility (distinct semantic category count), originality (statistical infrequency relative to a population), novelty (Likert-scale ratings by human judges), and other metrics~\citep{dumas2014understanding,runco1992scoring}.

Several computational methods have been proposed to automate these scores. Unsupervised approaches estimate (i)~flexibility by measuring the semantic diversity of an idea set~\citep{snyder2004creativity,bossomaier2009semantic}, and (ii) human judges' novelty ratings by computing an idea’s semantic distance from the task prompt~\citep{beaty2021automating,dumas2021measuring,acar2014assessing}. Hybrid and supervised methods directly predict novelty ratings using regression and clustering-based pipelines~\citep{organisciak2023beyond,stevenson2020automated}. However, these methods face generalizability issues, with models trained on one task or dataset often performing poorly on another~\citep{buczak2023machines}. More recently, studies have explored LLMs for zero-shot creativity scoring, but results show little to no correlation with human labels~\citep{chakrabarty2024art}.

Unfortunately, computational approaches for scoring ideas by \textit{statistical rarity} remain underexplored. Recent work has addressed related challenges---for example, \citet{lu2024ai} contrast AI model outputs against an extrinsic human-generated text corpora using $n$-gram overlap and Word Mover’s Distance (WMD) to probe the origin of AI creativity. However, using purely lexical ($n$-gram) or embedding-based matching methods (WMD) for operationalizing social comparison holds the risk of conflating distinct intents or over-separating true rephrasings, as they privilege surface similarity over conceptual intent~\citep{olson2021naming}. Our approach addresses these limitations by incorporating a zero-shot LLM in the annotation loop to make the \textit{subjective}, \textit{intent-sensitive} bucketing judgments, which, in turn, enables frequency-based originality scoring at scale.

\subsection{Text Clustering and Annotation}

LLMs have recently been explored for zero- and few-shot text clustering and annotation~\citep{xiao2023supporting}. Deductive clustering methods prompt LLMs to partition a given set of texts, generating categories or groupings directly~\citep{Viswanathan2024,chew2023llm}. However, most LLM-based deductive clustering methods assume all clusters are discoverable upfront and perform poorly when the concept space evolves. Inductive annotation methods, on the other hand, present labeled exemplars to classify new instances incrementally~\citep{dai2023llm}. While current approaches show promise on well-bounded tasks like topic labeling or thematic analysis, it remains unclear how best to navigate fat-tailed distributed datasets, where cluster (i.e., idea bucket) counts grow without bound as data scale increases.

\begin{table*}[!htbp]
\centering
  \begin{tabular}{lcccc}
    \hline
    \textbf{Dataset} & \textbf{\# Participants}& \textbf{\# Tasks} & \textbf{\# Ideas} & \textbf{\# Judges}\\
    \hline
    \texttt{socialmuse24}~\citep{baten2024ai}       & $109$    & $5$ & $5703$ &  $2$ \\
    \texttt{beaty18}~\citep{beaty2018robust}    & $171$      & $2$ & $2917$  & $4$\\
   \texttt{silvia17}~\citep{silvia2017old}    & $141$      & $2$ & $2355$  &$3$\\
   \texttt{beaty12}~\citep{beaty2012ideas}    & $133$      & $1$ & $1807$  &$3$\\
   \texttt{mohr16}~\citep{hofelich2016thinking}    & $305+284$      & $1+1$ & $1930+1582$   &$4$\\
  \hline
\end{tabular}
\caption{\label{dataset}
     Dataset summary. Each participant did one task in \texttt{mohr16}. In other datasets, all participants did all tasks.
  }
\end{table*}

\subsection{LLM-as-a-Judge}

The LLM-as-a-judge paradigm has emerged as a powerful approach for evaluating, ranking, and filtering outputs across tasks such as summarization, translation, alignment, and reasoning in NLP~\citep{li2024generation,liang2023encouraging,zhao2024diffagent}. Unlike earlier evaluation approaches~\citep{papineni2002bleu,zhang2019bertscore}, judge LLMs can assess contextual fit, intent, and subtle distinctions between candidates, using pointwise, pairwise, or listwise formats~\citep{gao2023human,shen2024comparative}.

Our task combines listwise judgment with decision-making: the LLM determines whether a new idea matches any retrieved exemplar or forms a new semantic bucket, akin to selection-based judgment~\citep{li2024dalk,yao2023tree}. We adopt a modular retrieval-based framework~\citep{lewis2021retrieval,izacard2020distilling}, where retrieval and codebook management are handled externally~\citep{khandelwal2020generalization}, leaving the stateless LLM to focus on subjective bucketing decisions. This separation improves stability and preserves the interpretability and psychometric auditability critical for creativity research.

\section{Dataset Acquisition}

We use five Alternative Uses Test datasets (Table~\ref{dataset}). Each dataset includes one or more tasks where participants generate alternative uses for an everyday object. Responses range from short phrases to full sentences (e.g., for a shoe: ``We can use a shoe as a hamster bed'' or ``As a doorstop'').

\subsection{Primary Dataset: \texttt{socialmuse24}}

We use the \texttt{socialmuse24} dataset to establish criterion validity~\citep{baten2024ai}. Two trained research assistants (H1 and H2) independently bucketed rephrased ideas within each task. The annotators saw the ideas in a random order. They followed the coding rules of Bouchard and Hare~\citep{bouchard1970size} and Guilford’s scoring key~\citep{guildford1978alternate}. Each idea thus has two categorical bucket IDs---one per annotator---serving as ground truth for evaluating our method. In the original study, these human-assigned buckets were used to estimate originality without automation; our goal is to replicate this process computationally. The dataset also includes flexibility-based \textit{Creativity Quotient} scores~\citep{snyder2004creativity,bossomaier2009semantic}, which we use to assess convergent validity.

\subsection{Secondary Datasets}

We draw on four publicly available AUT datasets to assess convergent and external validity~\citep{organisciak2023beyond,beaty2021automating}. Unlike \texttt{socialmuse24}, these datasets lack human bucketing annotations and thus cannot be used for our primary originality-scoring goal.

The \texttt{beaty18} dataset~\citep{beaty2018robust} includes four judges’ \textit{Creative Quality} ratings on a 1–5 scale, along with measures of: (i)~\textit{Creative Metaphor Generation}, where participants produced novel metaphors for two open-ended prompts, each rated 1–5 by four judges~\citep{beaty2013metaphorically}; (ii)~\textit{Big Five Personality}, assessing the participants' neuroticism, extraversion, openness to experience, agreeableness, and conscientiousness via standard questionnaires~\citep{mccrae2005neo}; (iii)~\textit{Fluid Intelligence}, through sequence-completion tasks with images~\citep{cattell1960measuring}, letters~\citep{ekstrom1976manual}, and numbers~\citep{Ces_1938}; and (iv)~\textit{Creative Self-concept}, via self-efficacy and self-identity questionnaires~\citep{karwowski2014creative}.

The \texttt{silvia17} dataset~\citep{silvia2017old} provides three judges’ \textit{Creative Quality} ratings, as well as openness-to-experience \textit{Personality} scores~\citep{lee2004psychometric}. The \texttt{beaty12} dataset~\citep{beaty2012ideas} includes three judges’ \textit{Creative Quality} ratings, plus \textit{Big Five Personality}, \textit{Creative Metaphor Generation}, and \textit{Fluid Intelligence} measures, paralleling \texttt{beaty18}.

Finally, \texttt{mohr16}~\citep{hofelich2016thinking} contains four judges’ ratings of idea \textit{Originality} and \textit{Flexibility}. Here, originality captured the uncommonness, remoteness, and cleverness of responses (1–5 scale)~\citep{silvia2008assessing}, while flexibility was defined as the number of categories in each participant’s responses, averaged across judges.

\section{Task Description}

\subsection{Problem Formulation}

Let \( \mathcal{P} = \{p_1, p_2, \ldots, p_N\} \) denote a corpus of \( N \) participants, each completing \( T \) ideation tasks. For each task \( t \in \{1, \ldots, T\} \), participant \( p_i \) produces a variable-length set of \( n_{i,t} \) free-form textual responses, denoted \( \mathcal{I}_{i,t} = \{x_{i,t}^{(1)}, \ldots, x_{i,t}^{(n_{i,t})} \} \).

Let \( \mathcal{X}_t = \bigcup_{i=1}^{N} \mathcal{I}_{i,t} \) denote the full idea set for task \( t \). The goal is to induce a task-specific partition \( \mathcal{B}_t = \{B_{t,1}, \ldots, B_{t,K_t}\} \) over \( \mathcal{X}_t \), where each `bucket' \( B_{t,k} \subseteq \mathcal{X}_t \) contains semantically equivalent ideas expressing the same underlying concept. 

Let \( k(x) \) denote the index of the bucket to which idea \( x \in \mathcal{X}_t \) is assigned. We define \( m_{t,k} \) as the number of distinct participants contributing at least one idea to bucket \( B_{t,k} \). Importantly, the bucketing is performed \textit{within} each task and \textit{across} participants, and no bucket identity is shared across tasks.

\subsection{Originality Metrics}

We explore $4$ frequency-based originality metrics:

(i)~\textbf{\texttt{rarity}}: Each idea bucket $B_{t,k}$ is scored as $(1 - \tfrac{m_{t,k}}{N})$, reflecting the bucket's relative infrequency in the sample~\citep{forthmann2020scrutinizing,forthmann2017typing}. A participant’s unnormalized \texttt{rarity} score is the sum across their ideas: $R_{i,t}^{\texttt{rarity}} = \sum_{x \in \mathcal{I}_{i,t}} \Big(1 - \tfrac{m_{t,k(x)}}{N}\Big)$.

(ii)~\textbf{\texttt{shapley}}: Each bucket $B_{t,k}$ is scored as $\tfrac{1}{m_{t,k}}$, making a bucket’s marginal value inversely proportional to the number of participants sharing it~\citep{page2018model}. A participant’s unnormalized \texttt{shapley} score is the sum across their ideas: $R_{i,t}^{\texttt{shapley}} = \sum_{x \in \mathcal{I}_{i,t}} \tfrac{1}{m_{t,k(x)}}$.

(iii)~\textbf{\texttt{uniqueness}}: Ideas in singleton buckets ($m_{t,k}=1$) receive a score of 1, and all others receive 0~\citep{forthmann2020scrutinizing,baten2021cues,baten2024ai}. A participant’s unnormalized \texttt{uniqueness} score is the count of their unique ideas:
$R_{i,t}^{\texttt{uniqueness}} = \sum_{x \in \mathcal{I}_{i,t}} \mathbb{I}\{m_{t,k(x)} = 1\}$.

(iv)~\textbf{\texttt{threshold}}: Ideas are scored by a tiered function $S(x)$ based on bucket prevalence~\citep{olson2021naming,deyoung2008cognitive,forthmann2020scrutinizing}:  
\[
S(x) =
\begin{cases}
3 & \text{if } \tfrac{m_{t,k(x)}}{N} \leq 0.01, \\
2 & \text{if } 0.01 < \tfrac{m_{t,k(x)}}{N} \leq 0.03, \\
1 & \text{if } 0.03 < \tfrac{m_{t,k(x)}}{N} \leq 0.10, \\
0 & \text{otherwise}.
\end{cases}
\]
A participant’s unnormalized \texttt{threshold} score is the sum of these scores:
$R_{i,t}^{\texttt{thresh}} = \sum_{x \in \mathcal{I}_{i,t}} S(x)$.

To compute a participant's overall unnormalized score across all tasks, we take $R_i^{\texttt{metric}} = \sum_{t=1}^{T} R_{i,t}^{\texttt{metric}}$. To account for fluency (i.e., the number of ideas, \( n_{i,t} \), contributed by participant \( p_i \) in task \( t \)), we define normalized originality as, $O_{i,t}^{\texttt{metric}} = \frac{R_{i,t}^{\texttt{metric}}}{n_{i,t}}$ and $O_i^{\texttt{metric}} = \sum_{t=1}^{T} O_{i,t}^{\texttt{metric}}$. 

\subsection{Evaluation Strategy}

We assess construct validity along two dimensions: (i)~alignment between computational and human idea-to-bucket clustering, and (ii)~agreement in participant-level originality scoring.

\textbf{Bucket-level construct validity.} The bucket labels are categorical and arbitrary. Moreover, the bucket sizes follow a fat-tailed distribution with a few highly frequent buckets and many rare ones (see \S\ref{understanding-human-dist}). Thus, traditional clustering metrics (e.g., Adjusted Rand Index) can be misleading due to being inflated by rare buckets. We therefore adopt \textit{Adjusted Mutual Information (AMI)}~\citep{Vinh2010} as our primary metric to evaluate idea-to-bucket clustering alignment between our proposed method and human annotations. This metric adjusts for chance agreement, is robust to label permutation and skewed distributions, and is well-suited for comparing clusterings with different numbers of clusters. For insight development, we also use \textit{Normalized Mutual Information (NMI)}~\citep{Vinh2010}, which quantifies mutual dependence between clusterings without chance correction, and \textit{V-measure}~\citep{rosenberg2007v}, which is the harmonic mean of homogeneity and completeness, reflecting both internal purity and cross-cluster coverage.

\textbf{Participant-level construct validity.} For originality scoring agreement, we use (i)~\textit{Zero-order Correlations} (Pearson's $r$ for linear agreement and Spearman’s $\rho$ for monotonic consistency), (ii)~\textit{Intraclass Correlation Coefficient} for consistency across judges~\citep{shrout1979intraclass}, and (iii)~\textit{Bland--Altman Plots} to identify systematic, scale-level biases~\citep{bland1986statistical}.

\textbf{Convergent and external validity.} \textit{Convergent validity} is assessed by correlating model originality scores with theoretically aligned creativity metrics (e.g., Creativity Quotient and Creative Quality Ratings). \textit{External validity} is evaluated by correlating model scores with established psychological and cognitive variables: personality traits, creative metaphor generation ability, fluid intelligence, and creative self-concept~\citep{beaty2021automating}.

\section{Understanding Human-Annotated Ground Truth Characteristics}\label{understanding-human}
\subsection{Distributional Properties of Idea Buckets}\label{understanding-human-dist} 
We assess the structure of idea diversity in \texttt{socialmuse24} using H1 and H2's buckets. H1 created more buckets per task ($399.6$, $95$\% CI: $[354.1, 445.1]$) than H2 ($230.8$ $[192.8, 268.8]$), indicating finer-grained distinctions.

\begin{figure}[t]
\centering
  \includegraphics[width=0.9\linewidth]{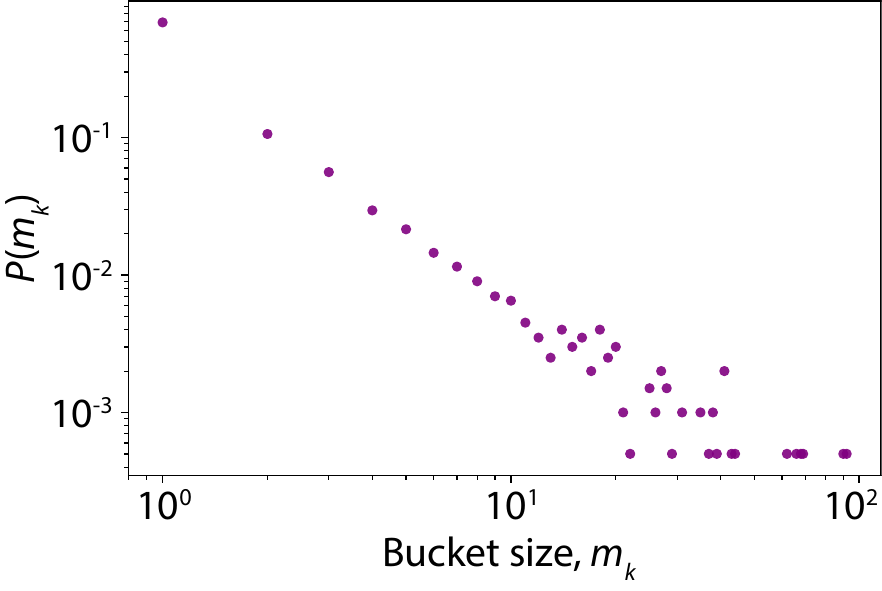}
  \caption {Idea bucket size distribution based on annotator H1's bucketing. See Figure~\ref{loglogkrish} for H2's case.}\label{loglogali}
\end{figure}

To examine bucket size (idea frequency) distributions, we fit a discrete power-law model to the bucket sizes for each task and compare it to a lognormal distribution via a likelihood ratio test~\citep{clauset2009power}. Both annotators produced fat-tailed distributions, with scaling exponents \( \alpha_{\text{H1}} = 2.01 \) $[1.73, 2.28]$ and \( \alpha_{\text{H2}} = 1.74 \) $[1.60, 1.88]$, consistent with power-law like behavior in linguistic and social systems ($\alpha\approx2$ to $3$)~\citep{newman2018networks}. This confirms that a few buckets are highly frequent while many are rare (Figure~\ref{loglogali}). However, the power-law model is not statistically favored over lognormal (\( P \geq 0.05 \)), suggesting that despite being fat-tailed, bucket size distributions are not strictly power-law and may be better described by lognormal or other alternatives.

\subsection{Inter Human Annotator Agreement on Idea-level Bucketing}
H1 and H2 show a mean AMI of $0.66$ $[0.64, 0.68]$, indicating strong alignment beyond what would be expected by random bucketing. NMI elucidates how informative one annotator’s bucketing is about the other's without adjusting for chance (i.e., NMI is less conservative). As expected, the mean NMI is higher at $0.85$ $[0.84, 0.87]$, reflecting strong underlying structure shared across annotators (Table~\ref{inter-human-cl}).

V-measure also yields a high mean of $0.85$ $[0.84, 0.87]$. Its homogeneity component ($0.80$) shows that H1’s buckets are reasonably pure with respect to H2, and its high completeness component ($0.92$) shows that H2's buckets almost perfectly recover H1's buckets. This pattern corroborates that H1 split buckets more finely than H2, but both annotators identified similar idea groupings. 

Overall, the annotators strongly agreed on their idea bucketing, despite granularity differences.

\subsection{Inter Human Annotator Agreement on Participant-level Originality Scoring}
We compute participant-level $\{O_i^{\texttt{metric}}\}$ using H1 and H2's bucket assignments and assess agreement.
 
The \texttt{threshold} and \texttt{shapley} metrics show the strongest correlations (\texttt{threshold}: $r = 0.77$ $[0.69, 0.84]$; \texttt{shapley}: $r = 0.79$ $[0.70, 0.85]$;  both $P < 0.001$). \texttt{uniqueness} and \texttt{rarity} show lower but still good correlations (\texttt{uniqueness}: $r = 0.73$ $[0.63, 0.81]$; \texttt{rarity}: \( r = 0.72 \) $ [0.61, 0.81]$; both \( P < 0.001 \); see Table~\ref{inter-human-corr} for $\rho$ estimates). 

The \texttt{threshold} and \texttt{shapley} metrics also show the strongest average consistency across judges: \( ICC(3,k) = 0.85 \) $[0.78, 0.90]$, \( P < 0.001 \) for both. \texttt{uniqueness} yields the lowest but good agreement: \( ICC(3,k) = 0.8 \) $[0.71, 0.86]$, \( P < 0.001 \) (Table~\ref{inter-human-icc}). Together, we note strong agreements in originality scoring across the human annotators.

\subsection{Insights for \textsc{MuseScorer} Development}\label{design-impl-MuseScorer}

These analyses help establish expectations for machine-based originality scoring. \textit{First}, human-annotated bucket sizes exhibit a fat-tailed structure. Any automated scoring system must account for this characteristic for its bucketing performance to approach the strong AMI baseline of humans.

\textit{Second}, based on the above evidence, we take the \texttt{threshold}-based normalized scores, $\{O_i^{\texttt{thresh}}\}$, as our person-level gold standard against which we evaluate machine-based originality scoring. We test for robustness against the other metrics.

\section{The \textsc{MuseScorer} System}
\subsection{Insights from Early Prototypes}

Our first prototype mimicked a human annotator's workflow by comparing each new idea against an expanding codebook of \textit{all} prior buckets. However, LLM prompts became intractable when $K_t$ exceeded roughly $150$. Given the scale-free bucket size distributions, massive corpora can have very large $K_t$, making exhaustive prompting infeasible. We therefore shifted to a retrieval-based approach, selecting a small candidate set for LLM judgment.

A second prototype let the LLM handle retrieval, decision-making, and codebook updates end-to-end, but this proved brittle---especially with smaller models (e.g., \texttt{phi4}). To improve stability, we offloaded retrieval and codebook management to external modules, leaving the LLM to focus solely on subjective bucketing.

\begin{algorithm}[t]
\caption{\textsc{MuseScorer}: LLM-Based Incremental Bucketing for a Single Creativity Task}\label{alg}
\begin{algorithmic}[1]
\Require Idea set \( \mathcal{X} = \{x_1, x_2, \ldots, x_{|\mathcal{X}|}\} \), LLM, candidate dictionary size \( K_c \)
\Ensure Partition \( \mathcal{B} = \{B_1, \ldots, B_K\} \), assignment map \( k(x) \)

\State Initialize empty codebook \( \mathcal{C} \gets \emptyset \)
\State Initialize bucket index \( K \gets 0 \)

\ForAll{ideas \( x \in \mathcal{X} \)}
    \If{\( |\mathcal{C}| \leq K_c \)}
        \State \( \mathcal{D}_x \gets \mathcal{C} \)
    \Else
        \State Use $K$-NN search to find top-\( K_c \) closest entries in \( \mathcal{C} \) to \( x \)
        \State \( \mathcal{D}_x \gets \{(k_j, d_j)\}_{j=1}^{K_c} \)
    \EndIf

    \State Query LLM: ``Is \( x \) a rephrasing of any \( d_j \in \mathcal{D}_x \)? Return \( k_j \) or $-1$.'' (In CoT prompting, also return a justification sentence)

    \If{LLM returns \( k^* \neq -1 \)}
        \State Assign \( k(x) \gets k^* \)
        \State \( B_{k^*} \gets B_{k^*} \cup \{x\} \)
    \Else
        \State \( K \gets K + 1 \)
        \State Create new bucket \( B_K \gets \{x\} \)
        \State Update codebook \( \mathcal{C} \gets \mathcal{C} \cup \{(K, x)\} \)
        \State Assign \( k(x) \gets K \)
    \EndIf
\EndFor

\State \Return \( \mathcal{B} = \{B_1, \ldots, B_K\} \), \( k(x) \) $\forall$ \( x \in \mathcal{X} \)
\end{algorithmic}
\end{algorithm}

\subsection{\textsc{MuseScorer} System Architecture}
Algorithm~\ref{alg} summarizes \textsc{MuseScorer}'s workflow. The LLM processes one idea at a time and assigns it to a semantically equivalent bucket or creates a new one. A dynamic codebook is initialized for each ideation task and updated as new ideas arrive. For each idea \( x \in \mathcal{X} \), a dictionary of candidate buckets \( \mathcal{D}_x \) is constructed via $K$-NN-based semantic search over the current codebook~\citep{khandelwal2020generalization}. $\mathcal{D}_x$ has a maximum size of $K_c$. When the number of existing buckets is smaller than \( K_c \), all of those buckets are taken in \( \mathcal{D}_x \). Each candidate in the dictionary \( \mathcal{D}_x = \{(k_j, d_j)\}_{j=1}^{K_c} \) maps bucket IDs to representative descriptions. 

We employ two kinds of prompting strategies: (i)~In \texttt{vanilla} prompting, the LLM determines whether \( x \) is a rephrasing of any \( d_j \). If so, it returns the corresponding key \( k_j \); otherwise, it returns \(-1\), signaling the creation of a new bucket with \( x \) as its description. (ii)~In Chain-of-Thought (\texttt{CoT}) prompting, the LLM additionally provides a one-sentence reasoning~\citep{wei2022chain}. The codebook and bucket assignment are updated accordingly. 

It is important to distinguish this design from semantic-similarity-based clustering methods, which can be used to bucket ideas directly (we employ such methods as our computational baselines; see \S\ref{comp_baselines}). Such methods typically attempt to assign ideas to clusters based on embedding distances, which risks conflating distinct intents or over-separating true rephrasings. In contrast, our use of $K$-NN-based retrieval only serves to keep the comparison space (and thereby prompt length) tractable; the final decision about bucket membership is made by the LLM. This separation ensures that semantic similarity supports efficiency, while the subjective, intent-sensitive aspect of bucketing remains with the judge LLM.

We fix \( K_c = 10 \) to allow a manageable prompt length while leaving sufficient margin for retrieval noise, and test robustness against other $K_c$ choices. We experiment with a factorial combination (`\textsc{MuseScorer} configurations') of LLM model variants (\texttt{llama3.3}, \texttt{qwen3}, and \texttt{phi4}), prompting strategies (\texttt{vanilla} and \texttt{CoT}), and sentence embeddings (Appendix Section~\ref{model_desc}).

\begin{table*}[ht]
\centering
\begin{tabular}{c l c c c c c}
\hline
 & \textbf{Model} & \textbf{AMI} & \textbf{NMI} & \textbf{Pearson's $r$} & \textbf{Spearman's $\rho$} & \textbf{ICC(3,1)} \\
\hline
\multirow{5}{*}{\rotatebox{90}{\textsc{MuseScorer}}} 
 & \texttt{llama3.3}, \texttt{CoT} & \textbf{0.59 ± 0.05} & \textbf{0.88 ± 0.02} & \textbf{0.88 ± 0.04} & \textbf{0.87 ± 0.05} & \textbf{0.88 ± 0.04} \\
 & \texttt{qwen3}, \texttt{CoT} & 0.56 ± 0.05 & 0.87 ± 0.02 & 0.79 ± 0.07 & 0.78 ± 0.07 & 0.77 ± 0.08 \\
 & \texttt{phi4}, \texttt{CoT} & 0.54 ± 0.01 & 0.83 ± 0.01 & 0.78 ± 0.08 & 0.76 ± 0.08 & 0.72 ± 0.09 \\
 & \texttt{llama3.3}, \texttt{vanilla} & 0.59 ± 0.03 & 0.86 ± 0.02 & 0.83 ± 0.06 & 0.79 ± 0.07 & 0.81 ± 0.06 \\
 & \texttt{phi4}, \texttt{vanilla} & 0.53 ± 0.02 & 0.83 ± 0.01 & 0.80 ± 0.07 & 0.78 ± 0.08 & 0.75 ± 0.08 \\
\hline
\multirow{4}{*}{\rotatebox{90}{Baseline}} 
 & $K$-means, Silhouette & 0.32 ± 0.09 & 0.86 ± 0.02 & 0.65 ± 0.11 & 0.67 ± 0.11 & 0.62 ± 0.12 \\
 & $K$-means, Semantic & 0.35 ± 0.06 & 0.87 ± 0.02 & 0.71 ± 0.10 & 0.70 ± 0.10 & 0.67 ± 0.10 \\
 & Aggl., Silhouette & 0.39 ± 0.02 & 0.85 ± 0.02 & 0.73 ± 0.09 & 0.68 ± 0.10 & 0.69 ± 0.10 \\
 & Aggl., Semantic & 0.31 ± 0.05 & 0.86 ± 0.02 & 0.65 ± 0.11 & 0.65 ± 0.11 & 0.61 ± 0.12 \\
\hline
\end{tabular}
\caption{Agreement metrics comparing computational models to H1's ground truths. Values are means ± half-width of the 95\% C.I. ($N = 109$). See Table~\ref{main_table_annotator2} for results based on H2's annotations, which replicate identical takeaways.}
\label{main_table}
\end{table*}

\section{Results and Discussion}
\subsection{Computational Baselines}\label{comp_baselines}
We use unsupervised clustering to establish a computational baseline for \textsc{MuseScorer}. We require algorithms that (i)~allow clusters of vastly different sizes, including fat-tail distributed ones, and (ii)~preserve singleton and rare buckets without dropping them as noise or outliers (\S\ref{design-impl-MuseScorer}). 

These constraints discourage us from using algorithms like DBScan (singleton and rare buckets are likely to be marked as noise)~\citep{ester1996dbscan} and HDBScan (minimum cluster size is $2$)~\citep{campello2013hdbscan}, and our experiments also corroborate their poor performance. $K$-means clustering is poor at handling imbalanced cluster sizes or shapes, and requires the number of clusters to be close to the number of datapoints to allow many singleton or rare buckets~\citep{macqueen1967kmeans}. Agglomerative hierarchical clustering is a reasonable choice for our constraints~\citep{ward1963hierarchical}. 

We report results with $K$-means and agglomerative algorithms. For each algorithm, we automatically search for the optimal number of buckets $K_t$ over the full range of $K_t=1$ to $|\mathcal{X}_t|$. To facilitate this search, we evaluate structural and semantic criteria using two metrics: (i)~\textit{Silhouette Score}, which assesses cluster quality based on geometric compactness and separation, with higher values indicating better-defined clusters~\citep{rousseeuw1987silhouettes}; and (ii)~\textit{Semantic Score}, which is the geometric mean of coherence (intra-cluster similarity) and exclusivity (inter-cluster distinctiveness), encouraging clusters that are both internally consistent and mutually distinct~\citep{mimno2011optimizing}.

\subsection{Distributional Properties of Computationally-labeled Idea Buckets}
We find that $K$-means and agglomerative algorithms produce an exorbitantly high $K_t$, with $831$ and $797$ buckets produced by the $K$-means algorithm (respectively based on Silhouette and Semantic scores), and $588$ and $838$ buckets by the agglomerative algorithm. For reference, $|\mathcal{X}_t|\approx 1141$ per task in \texttt{socialmuse24}. These bucket counts are significantly higher than H1 and H2's annotations ($P<0.001$; see \S\ref{understanding-human-dist}). In contrast, the \textsc{MuseScorer} configurations produce $K_t$ in the range of $255$ to $465$, overlapping those of the humans. The scaling exponents of $K$-means and agglomerative are systematically higher than the human baseline ($P<0.001$), but the \textsc{MuseScorer} configurations align with humans (Table~\ref{alpha_K_results}).

\subsection{Construct Validity of Idea-level Bucketing}
Table~\ref{main_table} and Figure~\ref{bucket} show the AMI and NMI agreements between H1 and machine bucketing. Taking H2 as the reference replicates identical insights (see Table~\ref{main_table_annotator2}). Interestingly, all methods score highly in the less conservative NMI metric and match the H1-H2 agreement, showing reasonable preservation of semantic grouping. 

However, when we correct for random chance and penalize mismatch in structure and granularity using the AMI metric, the \textsc{MuseScorer} configurations sustain human-like performance while the $K$-means and agglomerative algorithms suffer dramatically and systematically. Specifically, against a human-human AMI of $0.66$ $[0.64, 0.68]$, the \texttt{llama3.3} LLM with \texttt{CoT} prompting achieves the best AMI among the \textsc{MuseScorer} configurations at $0.59$ $[0.55,0.64]$, while the silhouette-tuned agglomerative algorithm manages the best AMI among the baseline models at a poor $0.39$ $[0.36,0.41]$. This is unsurprising, since a drop in AMI implies deviation from the structure and resolution of the human bucketing, which is corroborated by the systematically larger number of buckets $K$-means and agglomerative algorithms produce. In contrast, the \textsc{MuseScorer} configurations preserve more of the mutual structures, semantic coherence, and resolution, capturing up to $89\%$ of the fine-grained patterns humans see. 

Overall, \textsc{MuseScorer} shows strong idea-bucketing alignment with the humans, surpassing the performances of clustering-based baselines.

\begin{figure*}[t]
\centering
  \includegraphics[width=0.85\linewidth]{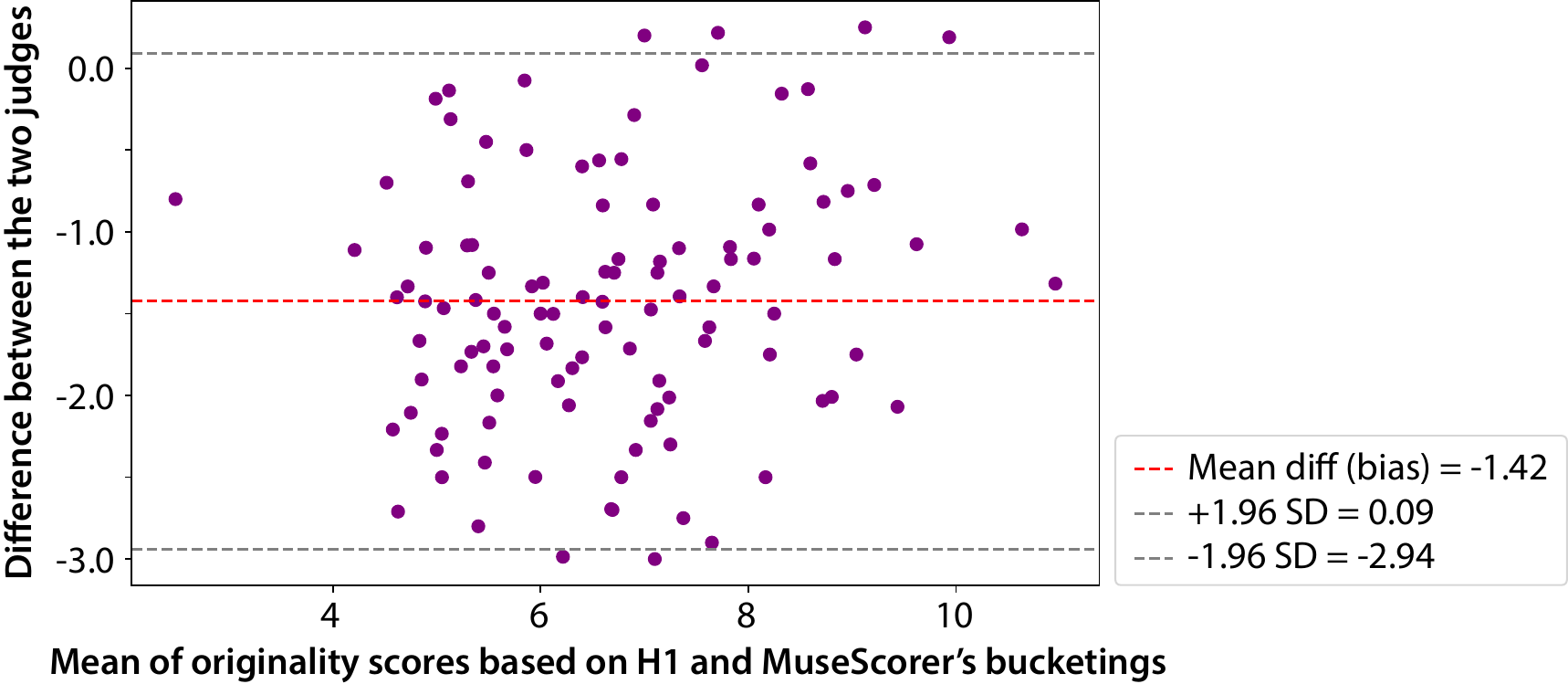}
  \caption {Bland-Altman visualization for bias detection.}\label{bland}
\end{figure*}

\subsection{Construct Validity of Participant-level Originality Scoring}
Table~\ref{main_table} and Figure~\ref{orig} show the participant-level $\{O_i^{\texttt{thresh}}\}$ score agreements based on H1 and machine bucketing. The results are robust to taking H2 as the reference (Table~\ref{main_table_annotator2}). \textsc{MuseScorer} with \texttt{llama3.3} and \texttt{CoT} prompting once again shows the best correlation (\( r = 0.89 \) $[0.83, 0.92]$, \( P < 0.001 \)). The baselines perform significantly worse, with the silhouette-tuned agglomerative algorithm achieving the best baseline correlation (\( r = 0.73 \) $[0.63, 0.81]$, \( P < 0.001 \)). 

\textsc{MuseScorer} with \texttt{llama3.3} and \texttt{CoT} prompting also shows the best $ICC(3,1) = 0.88$ $[0.83,0.92]$, \( P < 0.001 \). The clustering baselines reach a maximum of $ICC(3,1)=0.69$ $[0.57,0.77]$, \( P < 0.001 \), with the silhouette-tuned agglomerative model, performing significantly worse than \texttt{llama3.3} ($P<0.001$). Based on the above evidence, we pick \texttt{llama3.3} with \texttt{CoT} prompting as the default configuration for \textsc{MuseScorer} and use it for the remaining analysis.

We next visualize a Bland-Altman plot to identify systematic biases between H1 and \textsc{MuseScorer}-derived originality scores (Figure~\ref{bland}). $94.5\%$ of the points fall within the limits of agreement (LoA) of $\pm1.96$ SDs, and so does the mean difference (bias). This shows that \textsc{MuseScorer}-derived scores stay strongly in line with human scores across the originality spectrum. Although the proportional bias regression slope is slightly positive ($0.09)$, the effect is not statistically significant ($P > 0.05$), suggesting no systematic trend where the machine over- or under-scores ideas as originality increases. This supports the conclusion that \textsc{MuseScorer} provides stable, human-comparable originality assessments.

Taken together, \textsc{MuseScorer} shows strong construct validity in originality scoring against human ground truth. 

\subsection{Convergent and External Validity}

We evaluate \textsc{MuseScorer} for convergent and external validity against established creativity measures. Table~\ref{tab:validity} summarizes the correlations.

For convergent validity, \textsc{MuseScorer}'s normalized originality scores $\{O_i^{\texttt{thresh}}\}$ correlate strongly with Creativity Quotient (CQ) scores in \texttt{socialmuse24}. CQ is a flexibility measure that captures the diversity of semantic categories. However, CQ is unnormalized and confounded by idea fluency. Unsurprisingly, unnormalized $\{R_i^{\texttt{thresh}}\}$ scores show a stronger correlation with CQ. In rating-based datasets, \textsc{MuseScorer}'s $\{O_i^{\texttt{thresh}}\}$ scores correlate highly with human creative quality judgments (\texttt{beaty18}, \texttt{silvia17}, \texttt{beaty12}) and with rating-based originality in \texttt{mohr16}. The latter dataset also contains manually annotated flexibility scores, which do not account for fluency. Unsurprisingly, these flexibility scores correlate strongly with unnormalized $\{R_i^{\texttt{thresh}}\}$. Together, these findings confirm that \textsc{MuseScorer} captures core constructs of creativity with high fidelity.  

For external validity, \textsc{MuseScorer}'s $\{O_i^{\texttt{thresh}}\}$ scores correlate systematically with metaphor generation quality, openness to experience, and self-reported creative identity and self-efficacy. We did not observe systematic associations with fluid intelligence or other Big Five traits. Our results largely corroborate previous insights~\citep{beaty2021automating}, underscoring the system's broader external validity.

\begin{table*}[ht]
\centering
\begin{tabular}{l l c}
\hline
\textbf{Dataset } & \textbf{Comparison Variable} & \textbf{Correlation} \\
\hline
\multicolumn{3}{c}{\textit{Convergent Validity}} \\
\hline
\texttt{socialmuse24} & Creativity Quotient (CQ) & $r=0.40\,[0.23,0.55],\,P<0.001$ \\
\texttt{socialmuse24} & CQ vs. unnormalized $R_i^{\texttt{thresh}}$ & $r=0.48\,[0.32,0.62],\,P<0.001$ \\
\texttt{beaty18} & Creative Quality (mean ratings) & $r=0.77\,[0.71,0.83],\,P<0.001$ \\
\texttt{silvia17} & Creative Quality (mean ratings) & $r=0.54\,[0.41,0.65],\,P<0.001$ \\
\texttt{beaty12} & Creative Quality (mean ratings) & $r=0.42\,[0.27,0.55],\,P<0.001$ \\
\texttt{mohr16} & Rating-based Originality & $r=0.42\,[0.35,0.49],\,P<0.001$ \\
\texttt{mohr16} & Flexibility vs. unnormalized $R_i^{\texttt{thresh}}$ & $r=0.76\,[0.73,0.80],\,P<0.001$ \\
\hline
\multicolumn{3}{c}{\textit{External Validity}} \\
\hline
\texttt{beaty18} & Metaphor Generation & $r=0.17\,[0.02,0.32],\,P<0.05$ \\
\texttt{beaty12} & Metaphor Generation & $r=0.25\,[0.08,0.40],\,P<0.01$ \\
\texttt{beaty18} & Openness & $\rho=0.16\,[0.01,0.30],\,P<0.05$ \\
\texttt{beaty12} & Openness & $r=0.30\,[0.14,0.45],\,P<0.001$ \\
\texttt{silvia17} & Openness & $\rho=0.14\,[-0.02,0.30],\,P=0.09$ \\
\texttt{beaty18} & Creative Self-Identity & $r=0.34\,[0.19,0.48],\,P<0.001$ \\
\texttt{beaty18} & Creative Self-Efficacy & $r=0.29\,[0.14,0.44],\,P<0.001$ \\
\hline
\end{tabular}
\caption{Convergent and external validity of \textsc{MuseScorer}'s originality scores, $\{O_i^{\texttt{thresh}}\}$. Values are Pearson ($r$) or Spearman ($\rho$) correlations with $95$\% C.I. and significance levels.}\label{tab:validity}
\end{table*}

\subsection{Robustness}
The results depend on LLM, sentence embedding, and prompting strategy choices. We obtain the best \textsc{MuseScorer} results with a configuration comprising the \texttt{llama3.3:70b} LLM~\cite{llama3huggingface}, \texttt{e5-large-v2} sentence embedding~\cite{wang2022text}, and Chain-of-Thought prompting~\cite{wei2022chain} (\S\ref{model_desc}). We further probe this configuration's robustness across $K_c\in\{5,15\}$, and find results statistically similar to the default $K_c=10$. To assess ordering effects, we run the configuration with randomly ordered $\mathcal{X}_t$ across $3$ seeds. We find the results stable within the bounds reported in Table~\ref{main_table}. The main results with the \texttt{threshold} metric are largely reproduced by the other three metrics. But we find that \texttt{rarity} shows proportional bias in the Bland-Altman plot (slope = $0.2$, $P<0.01$), while \texttt{shapley} and \texttt{uniqueness} show no correlation with openness in the \texttt{silvia17} dataset, losing some external validity. The \texttt{threshold} metric thus emerges as the most robust choice for operationalizing statistical infrequency for originality scoring.

\section{Conclusion}

This work introduces \textsc{MuseScorer}, a scalable, zero-shot system for scoring the originality of creative ideas. By combining the LLM-as-a-judge paradigm with externally orchestrated retrieval, our method produces psychometrically aligned, intent-sensitive judgments without requiring task-specific fine-tuning or training data.

Across five distinct AUT datasets, \textsc{MuseScorer} demonstrates robust and consistent performance despite variation in task structures and idea distributions. Unlike opaque embedding-only approaches, our use of chain-of-thought (CoT) prompting yields interpretable outputs, allowing the system to provide justifications for bucketing decisions transparently.

Our approach is well-suited to support the growing body of research on human and AI creativity, particularly as large-scale, high-throughput studies become increasingly common~\citep{doshi2024generative,chakrabarty2025can,tanveer2018awe,kelty2023don}. By combining reliability, interpretability, and scale, this system expands the practical and methodological toolkit for researchers and opens new avenues for measuring and understanding creative potential in human and AI agents.

\section*{Limitations}

Several limitations should be considered in future work when extending frequency-based originality scoring. First, demographic fairness and accessibility remain important concerns. Variations in language use across cultural or educational backgrounds—especially in non-English contexts—may influence bucketing judgments and introduce bias if not carefully monitored.

Second, our validation is confined to AUT-style, text-based divergent thinking tasks. How well the approach generalizes to other creative domains (e.g., design, visual arts) remains an open question.

Third, while externally orchestrated retrieval mitigates some variability, the system remains sensitive to prompt length and phrasing~\citep{liu2023lost}. Subtle formatting changes can affect judgment quality, suggesting that prompt engineering and robustness testing deserve further study.

Fourth, efficiency may have room for improvement. We process ideas one at a time, which stabilizes performance—particularly for smaller models—but limits throughput. Future work could explore batching or multi-step reasoning to increase efficiency, though at potential cost to stability and computation.

Fifth, we kept the candidate retrieval size small ($K_c=\{5,10,15\}$). Larger candidate sets may improve coverage but increase token usage and cost. Similarly, our most effective \texttt{threshold} metric uses a heuristic tiering scheme adopted from prior literature; the robustness of these cutoffs remains to be validated.

Finally, as with all LLM-based systems, hallucination is a risk. In our case, hallucination manifests as misassigning an idea to the wrong bucket. However, \textsc{MuseScorer} achieves strong alignment with human annotations and passes psychometric validation despite this risk, suggesting the system is reasonably reliable within scope.

\section*{Ethical Considerations}
We reanalyzed public datasets from prior works (consistent with their intended use) and did not collect any new human data for this research. Given the nature of the research in creative assessment, we do not readily foresee potential harm or risk.

\clearpage
\appendix
\renewcommand{\thetable}{A\arabic{table}}
\setcounter{table}{0}
\renewcommand{\thefigure}{A\arabic{figure}}
\setcounter{figure}{0}

\section{Supplementary Materials}\label{sec:appendix}

\subsection{System Component Choices}\label{model_desc}

We experiment with the following system component alternatives: 

(i)~Large language models:  $\mathcal{M}$ = \{\texttt{llama3.3:70b-Instruct}~\citep{llama3huggingface,grattafiori2024llama3herdmodels}, \texttt{qwen3:32b}~\citep{yang2025qwen3technicalreport}, \texttt{phi4:14b}~\citep{abdin2024phi4technicalreport}\}. We pick these mid-sized, open-source models for their cost and computation efficiencies. 

(ii)~Sentence embedding models:  $\mathcal{E}$ = \{\texttt{all-mpnet-base-v2}~\citep{reimers2019sentencebert}, \texttt{bge-large-en-v1.5}~\citep{bge_embedding}, \texttt{e5-large-v2}~\citep{wang2022text}\}. These models are freely available on Huggingface and have been widely used in recent technological developments.

(iii)~Prompting strategies: $\mathcal{P}$ = \{\texttt{vanilla}, \texttt{CoT}~\citep{wei2022chain}\}.

In our experiments, we found the combination of \texttt{llama3.3:70b-Instruct}, \texttt{e5-large-v2}, and \texttt{CoT} to give the best performance.

\subsection{Experimentation Setup and GPU Usage}
We conducted all experiments using (i)~an Intel Core i7-based computer with $64$GB RAM and an RTX $3070$ Ti graphics card, and (ii) three MacBook Pro laptops. All our code and data are available on GitHub. The R\&D and final result generation took roughly $100$ GPU days.

\subsection{LLM Prompts}
\subsubsection*{System Prompt (Vanilla Prompting)}
\begin{promptbox}
You are an idea bucket annotator for ideas generated for the object \texttt{\{object\_name\}} in Guilford's Alternative Uses Test. You will be given an \texttt{input\_idea} to annotate against up to \texttt{\{comparison\_k\}} \texttt{comparison\_ideas}, given to you in a dictionary format with key-value pairs of \texttt{comparison\_idea\_ID: comparison\_idea\_description}. The keys are integers, and the values are strings. Your goal is to determine if the \texttt{input\_idea} is a very obviously rephrased version of one of those \texttt{comparison\_idea\_description}, or if it is slightly different.

\texttt{if input\_idea is a very obviously rephrased version of a certain comparison\_idea\_description:}\\
\phantom{123}\texttt{your\_annotation\_ID = comparison\_idea\_ID key of that comparison\_idea\_description value}

\texttt{elif input\_idea is a slightly different one:}\\
\phantom{123}\texttt{your\_annotation\_ID = -1}

Your response must be a text string containing exactly: \texttt{<your\_annotation\_ID>}.

For example: if your\_annotation\_ID is 6 since the input idea is a very obviously rephrased version of comparison\_idea\_ID 6, your response string should be "6". 
Another example: if your\_annotation\_ID is -1 because the input idea is not an obvious rephrasing of any comparison\_idea\_ID, your response string should be "-1".

Absolutely do not provide any extra text.
\end{promptbox}

\subsubsection*{System Prompt (CoT Prompting)}
\begin{promptbox}
You are an idea bucket annotator for ideas generated for the object \texttt{\{object\_name\}} in Guilford's Alternative Uses Test. You will be given an \texttt{input\_idea} to annotate against up to \texttt{\{comparison\_k\}} \texttt{comparison\_ideas}, given to you in a dictionary format with key-value pairs of \texttt{comparison\_idea\_ID: comparison\_idea\_description}. The keys are integers, and the values are strings. Your goal is to determine if the \texttt{input\_idea} is a very obviously rephrased version of one of those \texttt{comparison\_idea\_description}, or if it is slightly different.

\texttt{if input\_idea is a very obviously rephrased version of a certain comparison\_idea\_description:}\\
\phantom{123}\texttt{your\_annotation\_ID = comparison\_idea\_ID key of that comparison\_idea\_description value}

\texttt{elif input\_idea is a slightly different one:}\\
\phantom{123}\texttt{your\_annotation\_ID = -1}

You will also provide a \texttt{reason} string containing a single sentence explaining why you gave the \texttt{input\_idea} that specific \texttt{your\_annotation\_ID}.

Your response must be a text string containing exactly: \texttt{<your\_annotation\_ID><SPACE><reason>}.

For example: if your\_annotation\_ID is 6 and the reason is "The input idea is a very obviously rephrased version of comparison\_idea\_ID 6", your response string should be "6 The input idea is a very obviously rephrased version of comparison\_idea\_ID 6". 
Another example: if your\_annotation\_ID is -1 and the reason is "The input idea is not an obvious rephrasing of any comparison\_idea\_ID", your response string should be "-1 The input idea is not an obvious rephrasing of any comparison\_idea\_ID". 

Absolutely do not provide any extra text.
\end{promptbox}

\subsubsection*{User Prompt Per Idea (Both Conditions)}
\begin{promptbox}
input\_idea: \{idea\_text\}

comparison\_ideas: \{repr(comparison\_ideas)\}
\end{promptbox}

\subsection{AI Usage}
We used Grammarly AI to improve the grammatical accuracy of the manuscript, and ChatGPT to speed up the implementation of standard statistical analysis code. 

\clearpage
\subsection{Supplementary Tables and Figures}

\begin{table}[!htbp]
  \caption{\label{inter-human-cl}
     Inter-human annotator agreement on idea bucketing in \texttt{socialmuse24}.
  }
  \begin{tabular}{lc}
    \hline
    \textbf{Metric}           & \textbf{Mean} [$95$\% C.I.] \\
    \hline
    AMI      & $ 0.66$	$[0.64, 0.68]  $                                \\
    NMI    & 	$0.85$	$[0.84, 0.88]  $                                 \\
    V-measure       & $0.85$	$[0.84, 0.87]   $                                  \\
    Homogeneity & $0.80$	$[0.77, 0.82]  $                              \\
    Completeness   &                          	$0.92$	$[0.89, 0.95]   $      \\
    \hline
  \end{tabular}
\end{table}

\begin{table}[!htbp]
\caption{\label{inter-human-corr}
    Pearson and Spearman correlations between participant-level normalized $O^{\texttt{metric}}$ scores based on H1 and H2's bucketing. $N=109$ in all cases.
}
\begin{tabular}{llccc}
\hline
\textbf{Scoring Metric} & \textbf{Correlation Type} & \textbf{Estimate} & \textbf{$95\%$ C.I.} & \textbf{$P$-value} \\
\hline
\texttt{threshold}      & Pearson's $r$      & $0.77$ & $[0.69, 0.84]$ & $P < 0.001$ \\
                        & Spearman's $\rho$  & $0.75$ & $[0.65, 0.82]$ & $P < 0.001$ \\
\texttt{shapley}        & Pearson's $r$      & $0.79$ & $[0.70, 0.85]$ & $P < 0.001$ \\
                        & Spearman's $\rho$  & $0.74$ & $[0.64, 0.82]$ & $P < 0.001$ \\
\texttt{rarity}         & Pearson's $r$      & $0.72$ & $[0.61, 0.80]$ & $P < 0.001$ \\
                        & Spearman's $\rho$  & $0.64$ & $[0.51, 0.74]$ & $P < 0.001$ \\
\texttt{uniqueness}     & Pearson's $r$      & $0.73$ & $[0.63, 0.81]$ & $P < 0.001$ \\
                        & Spearman's $\rho$  & $0.66$ & $[0.54, 0.76]$ & $P < 0.001$ \\
\hline
\end{tabular}
\end{table}

\begin{table}[!htbp]
\caption{\label{inter-human-icc}ICC reliability of the participants' normalized originality $O^{\texttt{metric}}$ scores based on H1 and H2's bucketing.}
\begin{tabular}{lcccccc}
\hline
\textbf{Scoring Metric} & \textbf{$ICC(3,k)$} & \textbf{$F$} & \textbf{$df1$} & \textbf{$df2$} & \textbf{$P$-value} & \textbf{$95\%$ C.I.} \\
\hline
\texttt{threshold}     & $0.85$ & $6.79$ & $108$ & $108$ & $P < 0.001$ & $[0.78, 0.90]$ \\
\texttt{shapley}       & $0.85$ & $6.67$ & $108$ & $108$ & $P < 0.001$ & $[0.78, 0.90]$ \\
\texttt{rarity}        & $0.83$ & $5.73$ & $108$ & $108$ & $P < 0.001$ & $[0.75, 0.88]$ \\
\texttt{uniqueness}    & $0.80$ & $4.97$ & $108$ & $108$ & $P < 0.001$ & $[0.71, 0.86]$ \\
\hline
\end{tabular}
\end{table}

\begin{table}[!htbp]
\caption{Cluster count $K$ and power-law exponent $\alpha$ for various computational scoring methods.}
\label{alpha_K_results}
\begin{tabular}{lcc}
\hline
\textbf{Model} & \textbf{$K$ [$95\%$ C.I.]} & \textbf{$\alpha$ [$95\%$ C.I.]} \\
\hline
\texttt{llama3.3}, \texttt{CoT} & $465.4$ [$426.8$, $504.0$] & $2.28$ [$2.14$, $2.42$] \\
\texttt{qwen3}, \texttt{CoT} & $462.4$ [$432.7$, $492.1$] & $2.43$ [$2.20$, $2.67$] \\
\texttt{phi4}, \texttt{CoT} & $255.0$ [$207.3$, $302.7$] & $2.39$ [$1.72$, $3.05$] \\
\texttt{llama3.3}, \texttt{vanilla} & $367.8$ [$333.3$, $402.3$] & $2.29$ [$1.97$, $2.61$] \\
\texttt{phi4}, \texttt{vanilla} & $275.6$ [$229.5$, $321.7$] & $2.51$ [$2.23$, $2.78$] \\
\hline
$K$-means, Silhouette & $830.6$ [$729.2$, $932.0$] & $3.12$ [$2.82$, $3.43$] \\
$K$-means, Semantic & $797.4$ [$757.8$, $837.0$] & $3.12$ [$2.67$, $3.57$] \\
Agglomerative, Silhouette & $588.0$ [$524.9$, $651.1$] & $5.68$ [$1.26$, $10.09$] \\
Agglomerative, Semantic & $838.0$ [$815.9$, $860.1$] & $3.80$ [$2.63$, $4.97$] \\
\hline
\end{tabular}
\end{table}

\clearpage

\begin{table}[!htbp]
\caption{Agreement metrics comparing computational models to H2's ground truths. Values denote mean ± half-width of the 95\% C.I. ($N = 109$).}
\label{main_table_annotator2}
\begin{tabular}{c l c c c c c}
\hline
 & \textbf{Model} & \textbf{AMI} & \textbf{NMI} & \textbf{Pearson $r$} & \textbf{Spearman $\rho$} & \textbf{ICC(3,1)} \\
\hline
\multirow{5}{*}{\rotatebox{90}{\textsc{MuseScorer}}} 
 & \texttt{llama3.3}, \texttt{CoT} & 0.57 ± 0.04 & \textbf{0.84 ± 0.02} & \textbf{0.76 ± 0.08} & \textbf{0.74 ± 0.09} & 0.74 ± 0.09 \\
 & \texttt{qwen3}, \texttt{CoT} & 0.54 ± 0.04 & 0.83 ± 0.02 & 0.74 ± 0.09 & 0.73 ± 0.09 & 0.74 ± 0.09 \\
 & \texttt{phi4}, \texttt{CoT} & 0.56 ± 0.03 & 0.79 ± 0.01 & 0.67 ± 0.10 & 0.68 ± 0.10 & 0.67 ± 0.10 \\
 & \texttt{llama3.3}, \texttt{vanilla} & \textbf{0.59 ± 0.03} & 0.83 ± 0.01 & 0.76 ± 0.08 & 0.74 ± 0.09 & \textbf{0.75 ± 0.08} \\
 & \texttt{phi4}, \texttt{vanilla} & 0.55 ± 0.04 & 0.80 ± 0.01 & 0.73 ± 0.09 & 0.71 ± 0.10 & 0.73 ± 0.09 \\
\hline
\multirow{4}{*}{\rotatebox{90}{Baseline}} 
 & $K$-means, Silhouette & 0.28 ± 0.07 & 0.80 ± 0.02 & 0.59 ± 0.12 & 0.62 ± 0.12 & 0.59 ± 0.12 \\
 & $K$-means, Semantic & 0.30 ± 0.05 & 0.80 ± 0.02 & 0.66 ± 0.11 & 0.68 ± 0.10 & 0.66 ± 0.11 \\
 & Aggl., Silhouette & 0.36 ± 0.03 & 0.80 ± 0.02 & 0.65 ± 0.11 & 0.60 ± 0.12 & 0.64 ± 0.11 \\
 & Aggl., Semantic & 0.26 ± 0.05 & 0.80 ± 0.02 & 0.60 ± 0.12 & 0.64 ± 0.11 & 0.60 ± 0.12 \\
\hline
\end{tabular}
\end{table}

\clearpage

\begin{figure*}[t]
\centering
  \includegraphics[width=1\linewidth]{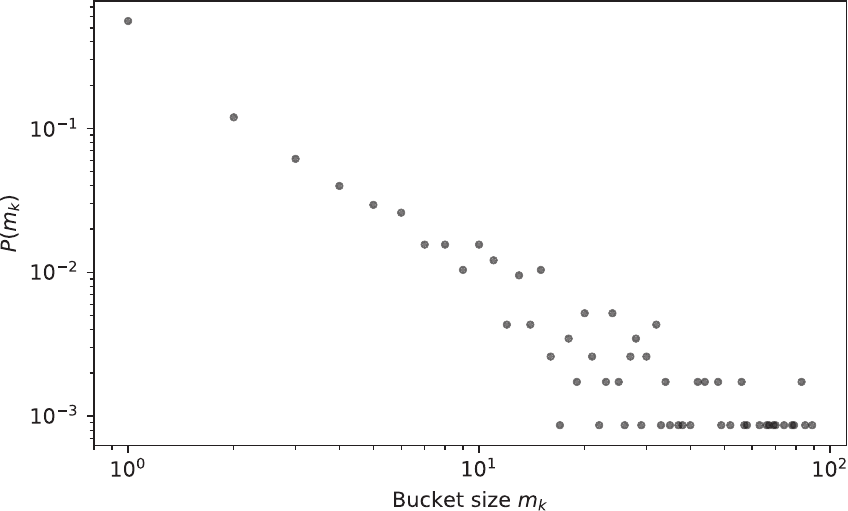}
  \caption {Idea bucket size distribution based on annotator H2's bucketing.}\label{loglogkrish}
\end{figure*}

\clearpage

\begin{figure*}[t]
\centering
  \includegraphics[width=1\linewidth]{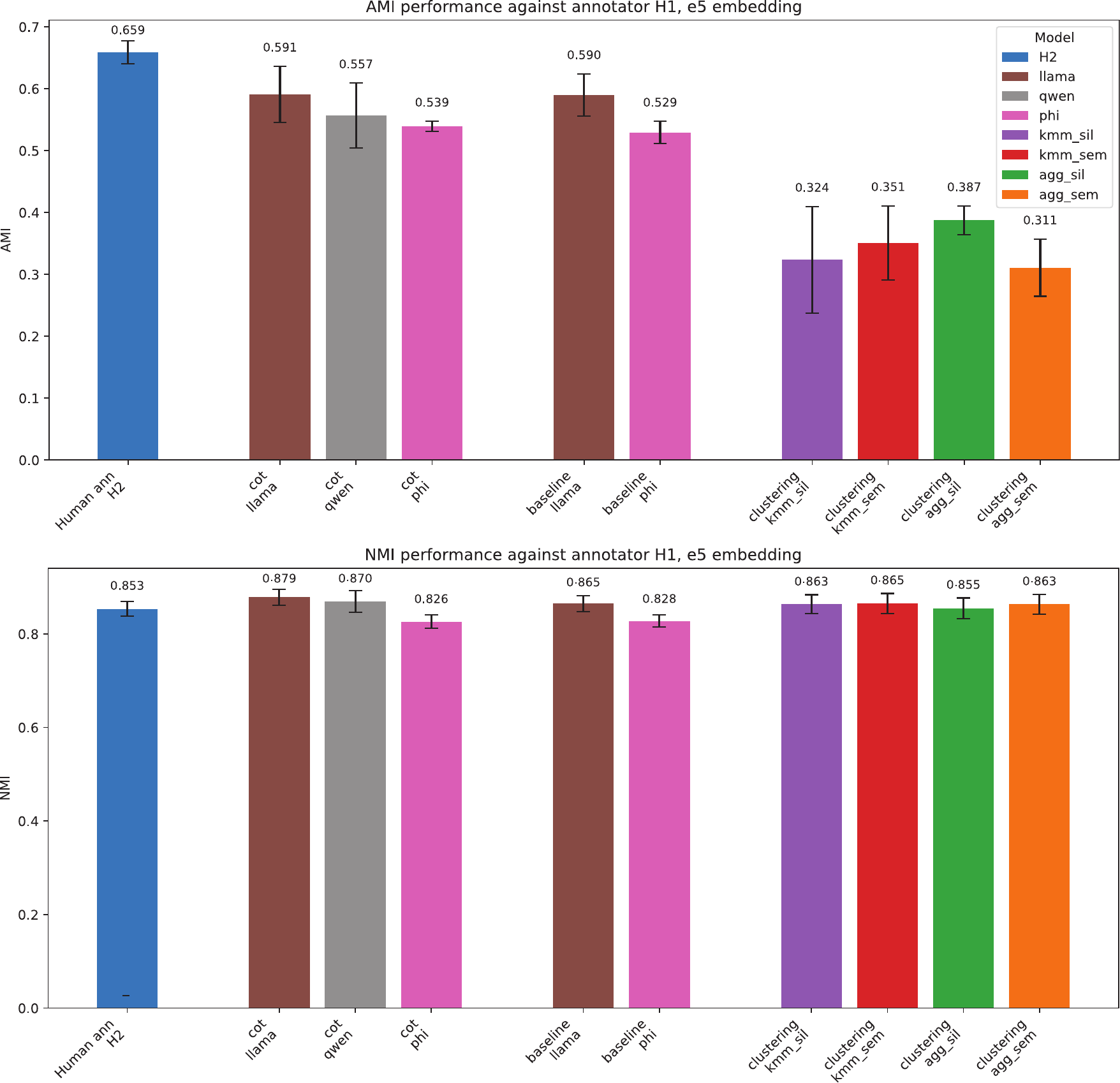}
  \caption {AMI and NMI performance comparison against annotator H1}\label{bucket}
\end{figure*}

\clearpage

\begin{figure*}[t]
\centering
  \includegraphics[width=1\linewidth]{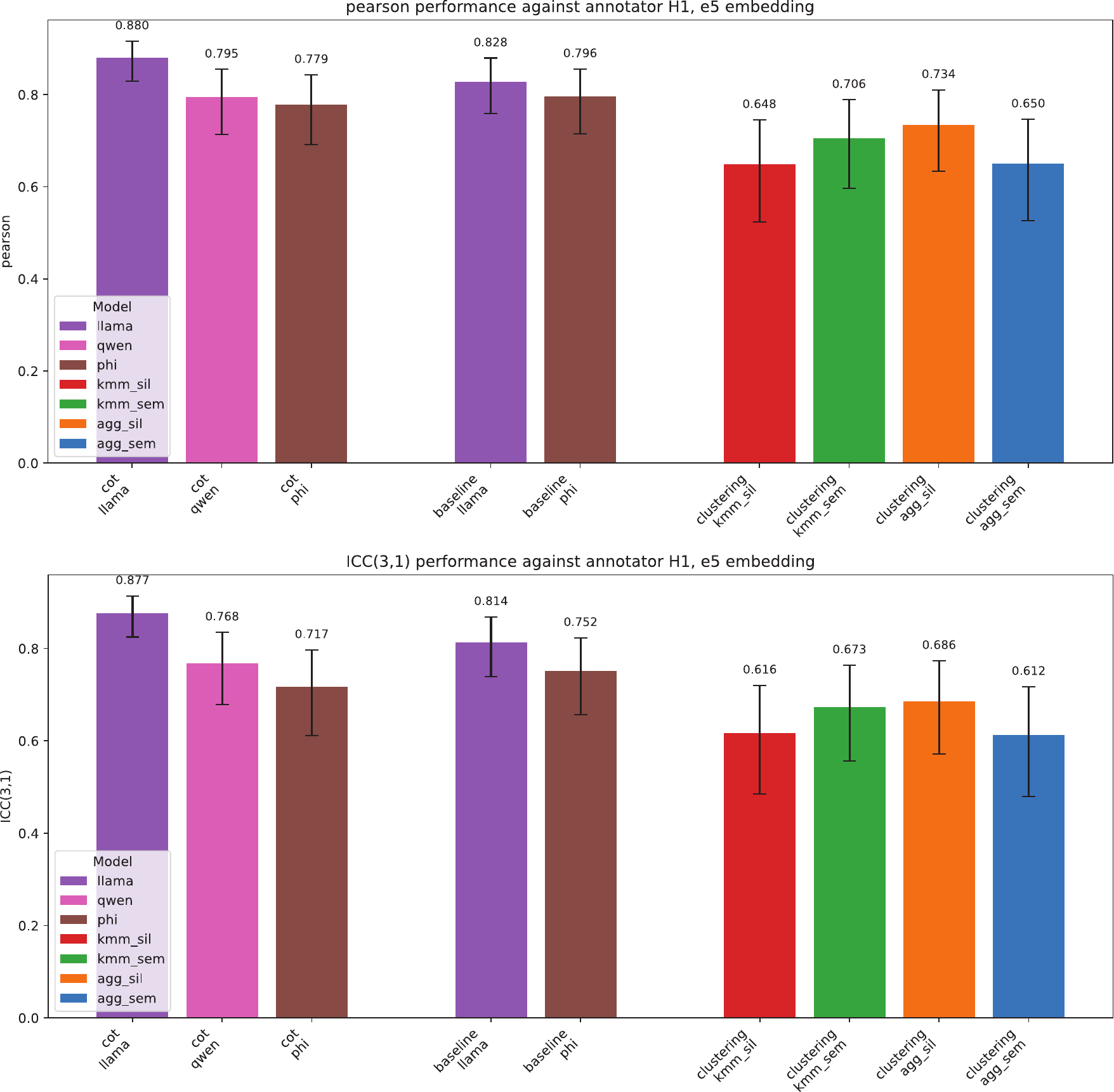}
  \caption {Pearson's $r$ and $ICC$ performance comparison against annotator H1}\label{orig}
\end{figure*}


\begin{thebibliography}{81}
\providecommand{\natexlab}[1]{#1}

\bibitem[{Abdin et~al.(2024)Abdin, Aneja, Behl, Bubeck, Eldan, Gunasekar, Harrison, Hewett, Javaheripi, Kauffmann, Lee, Lee, Li, Liu, Mendes, Nguyen, Price, de~Rosa, Saarikivi, Salim, Shah, Wang, Ward, Wu, Yu, Zhang, and Zhang}]{abdin2024phi4technicalreport}
Marah Abdin, Jyoti Aneja, Harkirat Behl, Sébastien Bubeck, Ronen Eldan, Suriya Gunasekar, Michael Harrison, Russell~J. Hewett, Mojan Javaheripi, Piero Kauffmann, James~R. Lee, Yin~Tat Lee, Yuanzhi Li, Weishung Liu, Caio C.~T. Mendes, Anh Nguyen, Eric Price, Gustavo de~Rosa, Olli Saarikivi, and 8 others. 2024.
\newblock \href {https://arxiv.org/abs/2412.08905} {Phi-4 technical report}.
\newblock \emph{Preprint}, arXiv:2412.08905.

\bibitem[{Acar and Runco(2014)}]{acar2014assessing}
Selcuk Acar and Mark~A Runco. 2014.
\newblock Assessing associative distance among ideas elicited by tests of divergent thinking.
\newblock \emph{Creativity Research Journal}, 26(2):229--238.

\bibitem[{Baten et~al.(2021)Baten, Aslin, Ghoshal, and Hoque}]{baten2021cues}
Raiyan~Abdul Baten, Richard~N Aslin, Gourab Ghoshal, and Ehsan Hoque. 2021.
\newblock Cues to gender and racial identity reduce creativity in diverse social networks.
\newblock \emph{Scientific Reports}, 11(1):10261.

\bibitem[{Baten et~al.(2022)Baten, Aslin, Ghoshal, and Hoque}]{baten2022novel}
Raiyan~Abdul Baten, Richard~N Aslin, Gourab Ghoshal, and Ehsan Hoque. 2022.
\newblock Novel idea generation in social networks is optimized by exposure to a ``{G}oldilocks''' level of idea-variability.
\newblock \emph{PNAS Nexus}, 1(5):pgac255.

\bibitem[{Baten et~al.(2020)Baten, Bagley, Tenesaca, Clark, Bagrow, Ghoshal, and Hoque}]{baten2020creativity}
Raiyan~Abdul Baten, Daryl Bagley, Ashely Tenesaca, Famous Clark, James~P Bagrow, Gourab Ghoshal, and Ehsan Hoque. 2020.
\newblock Creativity in temporal social networks: How divergent thinking is impacted by one's choice of peers.
\newblock \emph{Journal of the Royal Society Interface}, 17(171):20200667.

\bibitem[{Baten et~al.(2024)Baten, Bangash, Veera, Ghoshal, and Hoque}]{baten2024ai}
Raiyan~Abdul Baten, Ali~Sarosh Bangash, Krish Veera, Gourab Ghoshal, and Ehsan Hoque. 2024.
\newblock {AI} can enhance creativity in social networks.
\newblock \emph{arXiv preprint arXiv:2410.15264}.

\bibitem[{Beaty and Johnson(2021)}]{beaty2021automating}
Roger~E Beaty and Dan~R Johnson. 2021.
\newblock Automating creativity assessment with {S}em{D}is: An open platform for computing semantic distance.
\newblock \emph{Behavior Research Methods}, 53(2):757--780.

\bibitem[{Beaty et~al.(2018)Beaty, Kenett, Christensen, Rosenberg, Benedek, Chen, Fink, Qiu, Kwapil, Kane et~al.}]{beaty2018robust}
Roger~E Beaty, Yoed~N Kenett, Alexander~P Christensen, Monica~D Rosenberg, Mathias Benedek, Qunlin Chen, Andreas Fink, Jiang Qiu, Thomas~R Kwapil, Michael~J Kane, and 1 others. 2018.
\newblock Robust prediction of individual creative ability from brain functional connectivity.
\newblock \emph{Proceedings of the National Academy of Sciences}, 115(5):1087--1092.

\bibitem[{Beaty and Silvia(2012)}]{beaty2012ideas}
Roger~E Beaty and Paul~J Silvia. 2012.
\newblock Why do ideas get more creative across time? {A}n executive interpretation of the serial order effect in divergent thinking tasks.
\newblock \emph{Psychology of Aesthetics, Creativity, and the Arts}, 6(4):309.

\bibitem[{Beaty and Silvia(2013)}]{beaty2013metaphorically}
Roger~E Beaty and Paul~J Silvia. 2013.
\newblock Metaphorically speaking: Cognitive abilities and the production of figurative language.
\newblock \emph{Memory \& cognition}, 41:255--267.

\bibitem[{Beketayev and Runco(2016)}]{beketayev2016scoring}
Kenes Beketayev and Mark~A Runco. 2016.
\newblock Scoring divergent thinking tests by computer with a semantics-based algorithm.
\newblock \emph{Europe's Journal of Psychology}, 12(2):210.

\bibitem[{Bland and Altman(1986)}]{bland1986statistical}
J.~Martin Bland and Douglas~G. Altman. 1986.
\newblock \href {https://doi.org/10.1016/S0140-6736(86)90837-8} {Statistical methods for assessing agreement between two methods of clinical measurement}.
\newblock \emph{The Lancet}, 327(8476):307--310.

\bibitem[{Bossomaier et~al.(2009)Bossomaier, Harr{\'e}, Knittel, and Snyder}]{bossomaier2009semantic}
Terry Bossomaier, Mike Harr{\'e}, Anthony Knittel, and Allan Snyder. 2009.
\newblock A semantic network approach to the {C}reativity {Q}uotient ({CQ}).
\newblock \emph{Creativity Research Journal}, 21(1):64--71.

\bibitem[{Bouchard~Jr and Hare(1970)}]{bouchard1970size}
Thomas~J Bouchard~Jr and Melana Hare. 1970.
\newblock Size, performance, and potential in brainstorming groups.
\newblock \emph{Journal of Applied Psychology}, 54(1p1):51.

\bibitem[{Buczak et~al.(2023)Buczak, Huang, Forthmann, and Doebler}]{buczak2023machines}
Philip Buczak, He~Huang, Boris Forthmann, and Philipp Doebler. 2023.
\newblock The machines take over: A comparison of various supervised learning approaches for automated scoring of divergent thinking tasks.
\newblock \emph{The Journal of Creative Behavior}, 57(1):17--36.

\bibitem[{Campello et~al.(2013)Campello, Moulavi, and Sander}]{campello2013hdbscan}
Ricardo J. G.~B. Campello, Davoud Moulavi, and J\"org Sander. 2013.
\newblock Density-based clustering based on hierarchical density estimates.
\newblock In \emph{Pacific-Asia Conference on Knowledge Discovery and Data Mining (PAKDD)}, pages 160--172. Springer.

\bibitem[{Cattell and Cattell(1960)}]{cattell1960measuring}
Raymond~Bernard Cattell and Alberta~KS Cattell. 1960.
\newblock \emph{Measuring intelligence with the culture fair tests}.
\newblock Institute for Personality and Ability Testing.

\bibitem[{Chakrabarty et~al.(2024)Chakrabarty, Laban, Agarwal, Muresan, and Wu}]{chakrabarty2024art}
Tuhin Chakrabarty, Philippe Laban, Divyansh Agarwal, Smaranda Muresan, and Chien-Sheng Wu. 2024.
\newblock Art or artifice? {Large} language models and the false promise of creativity.
\newblock In \emph{Proceedings of the 2024 CHI Conference on Human Factors in Computing Systems}, pages 1--34.

\bibitem[{Chakrabarty et~al.(2025)Chakrabarty, Laban, and Wu}]{chakrabarty2025can}
Tuhin Chakrabarty, Philippe Laban, and Chien-Sheng Wu. 2025.
\newblock Can {AI} writing be salvaged? {M}itigating idiosyncrasies and improving human-{AI} alignment in the writing process through edits.
\newblock In \emph{Proceedings of the 2025 CHI Conference on Human Factors in Computing Systems}, pages 1--33.

\bibitem[{Chew et~al.(2023)Chew, Bollenbacher, Wenger, Speer, and Kim}]{chew2023llm}
Robert Chew, John Bollenbacher, Michael Wenger, Jessica Speer, and Annice Kim. 2023.
\newblock {LLM}-assisted content analysis: Using large language models to support deductive coding.
\newblock \emph{arXiv preprint arXiv:2306.14924}.

\bibitem[{Clauset et~al.(2009)Clauset, Shalizi, and Newman}]{clauset2009power}
Aaron Clauset, Cosma~Rohilla Shalizi, and Mark~EJ Newman. 2009.
\newblock Power-law distributions in empirical data.
\newblock \emph{SIAM Review}, 51(4):661--703.

\bibitem[{Dai et~al.(2023)Dai, Xiong, and Ku}]{dai2023llm}
Shih-Chieh Dai, Aiping Xiong, and Lun-Wei Ku. 2023.
\newblock {LLM}-in-the-loop: Leveraging large language model for thematic analysis.
\newblock \emph{arXiv preprint arXiv:2310.15100}.

\bibitem[{DeYoung et~al.(2008)DeYoung, Flanders, and Peterson}]{deyoung2008cognitive}
Colin~G DeYoung, Joseph~L Flanders, and Jordan~B Peterson. 2008.
\newblock Cognitive abilities involved in insight problem solving: An individual differences model.
\newblock \emph{Creativity Research Journal}, 20(3):278--290.

\bibitem[{Doshi and Hauser(2024)}]{doshi2024generative}
Anil~R Doshi and Oliver~P Hauser. 2024.
\newblock Generative {AI} enhances individual creativity but reduces the collective diversity of novel content.
\newblock \emph{Science Advances}, 10(28):eadn5290.

\bibitem[{Dumas and Dunbar(2014)}]{dumas2014understanding}
Denis Dumas and Kevin~N Dunbar. 2014.
\newblock Understanding fluency and originality: A latent variable perspective.
\newblock \emph{Thinking Skills and Creativity}, 14:56--67.

\bibitem[{Dumas et~al.(2021)Dumas, Organisciak, and Doherty}]{dumas2021measuring}
Denis Dumas, Peter Organisciak, and Michael Doherty. 2021.
\newblock Measuring divergent thinking originality with human raters and text-mining models: A psychometric comparison of methods.
\newblock \emph{Psychology of Aesthetics, Creativity, and the Arts}, 15(4):645.

\bibitem[{Ekstrom et~al.(1976)Ekstrom, French, Harman, and Dermen}]{ekstrom1976manual}
Ruth~B Ekstrom, John~W French, Harry~H Harman, and D~Dermen. 1976.
\newblock Manual for kit of factor-referenced tests.
\newblock \emph{Princeton, NJ: Educational Testing Service}, 586:1989--1995.

\bibitem[{Ester et~al.(1996)Ester, Kriegel, Sander, and Xu}]{ester1996dbscan}
Martin Ester, Hans-Peter Kriegel, J{\"o}rg Sander, and Xiaowei Xu. 1996.
\newblock A density-based algorithm for discovering clusters in large spatial databases with noise.
\newblock In \emph{Proceedings of the Second International Conference on Knowledge Discovery and Data Mining (KDD-96)}, pages 226--231.

\bibitem[{Forthmann et~al.(2017)Forthmann, Holling, {\c{C}}elik, Storme, and Lubart}]{forthmann2017typing}
Boris Forthmann, Heinz Holling, P{\i}nar {\c{C}}elik, Martin Storme, and Todd Lubart. 2017.
\newblock Typing speed as a confounding variable and the measurement of quality in divergent thinking.
\newblock \emph{Creativity Research Journal}, 29(3):257--269.

\bibitem[{Forthmann et~al.(2020)Forthmann, Paek, Dumas, Barbot, and Holling}]{forthmann2020scrutinizing}
Boris Forthmann, Sue~Hyeon Paek, Denis Dumas, Baptiste Barbot, and Heinz Holling. 2020.
\newblock Scrutinizing the basis of originality in divergent thinking tests: On the measurement precision of response propensity estimates.
\newblock \emph{British Journal of Educational Psychology}, 90(3):683--699.

\bibitem[{Gao et~al.(2023)Gao, Ruan, Sun, Yin, Yang, and Wan}]{gao2023human}
Mingqi Gao, Jie Ruan, Renliang Sun, Xunjian Yin, Shiping Yang, and Xiaojun Wan. 2023.
\newblock Human-like summarization evaluation with {ChatGPT}.
\newblock \emph{arXiv preprint arXiv:2304.02554}.

\bibitem[{Grattafiori et~al.(2024)Grattafiori, Dubey, Jauhri, Pandey, Kadian, Al-Dahle, Letman, Mathur, Schelten, Vaughan, Yang, Fan, Goyal, Hartshorn, Yang, Mitra, Sravankumar, Korenev, Hinsvark, Rao, Zhang, Rodriguez, Gregerson, Spataru, Roziere, Biron, Tang, Chern, Caucheteux, Nayak, Bi, Marra, McConnell, Keller, Touret, Wu, Wong, Ferrer, Nikolaidis, Allonsius, Song, Pintz, Livshits, Wyatt, Esiobu, Choudhary, Mahajan, Garcia-Olano, Perino, Hupkes, Lakomkin, AlBadawy, Lobanova, Dinan, Smith, Radenovic, Guzmán, Zhang, Synnaeve, Lee, Anderson, Thattai, Nail, Mialon, Pang, Cucurell, Nguyen, Korevaar, Xu, Touvron, Zarov, Ibarra, Kloumann, Misra, Evtimov, Zhang, Copet, Lee, Geffert, Vranes, Park, Mahadeokar, Shah, van~der Linde, Billock, Hong, Lee, Fu, Chi, Huang, Liu, Wang, Yu, Bitton, Spisak, Park, Rocca, Johnstun, Saxe, Jia, Alwala, Prasad, Upasani, Plawiak, Li, Heafield, Stone, El-Arini, Iyer, Malik, Chiu, Bhalla, Lakhotia, Rantala-Yeary, van~der Maaten, Chen, Tan, Jenkins, Martin, Madaan, Malo, Blecher,
  Landzaat, de~Oliveira, Muzzi, Pasupuleti, Singh, Paluri, Kardas, Tsimpoukelli, Oldham, Rita, Pavlova, Kambadur, Lewis, Si, Singh, Hassan, Goyal, Torabi, Bashlykov, Bogoychev, Chatterji, Zhang, Duchenne, Çelebi, Alrassy, Zhang, Li, Vasic, Weng, Bhargava, Dubal, Krishnan, Koura, Xu, He, Dong, Srinivasan, Ganapathy, Calderer, Cabral, Stojnic, Raileanu, Maheswari, Girdhar, Patel, Sauvestre, Polidoro, Sumbaly, Taylor, Silva, Hou, Wang, Hosseini, Chennabasappa, Singh, Bell, Kim, Edunov, Nie, Narang, Raparthy, Shen, Wan, Bhosale, Zhang, Vandenhende, Batra, Whitman, Sootla, Collot, Gururangan, Borodinsky, Herman, Fowler, Sheasha, Georgiou, Scialom, Speckbacher, Mihaylov, Xiao, Karn, Goswami, Gupta, Ramanathan, Kerkez, Gonguet, Do, Vogeti, Albiero, Petrovic, Chu, Xiong, Fu, Meers, Martinet, Wang, Wang, Tan, Xia, Xie, Jia, Wang, Goldschlag, Gaur, Babaei, Wen, Song, Zhang, Li, Mao, Coudert, Yan, Chen, Papakipos, Singh, Srivastava, Jain, Kelsey, Shajnfeld, Gangidi, Victoria, Goldstand, Menon, Sharma, Boesenberg,
  Baevski, Feinstein, Kallet, Sangani, Teo, Yunus, Lupu, Alvarado, Caples, Gu, Ho, Poulton, Ryan, Ramchandani, Dong, Franco, Goyal, Saraf, Chowdhury, Gabriel, Bharambe, Eisenman, Yazdan, James, Maurer, Leonhardi, Huang, Loyd, Paola, Paranjape, Liu, Wu, Ni, Hancock, Wasti, Spence, Stojkovic, Gamido, Montalvo, Parker, Burton, Mejia, Liu, Wang, Kim, Zhou, Hu, Chu, Cai, Tindal, Feichtenhofer, Gao, Civin, Beaty, Kreymer, Li, Adkins, Xu, Testuggine, David, Parikh, Liskovich, Foss, Wang, Le, Holland, Dowling, Jamil, Montgomery, Presani, Hahn, Wood, Le, Brinkman, Arcaute, Dunbar, Smothers, Sun, Kreuk, Tian, Kokkinos, Ozgenel, Caggioni, Kanayet, Seide, Florez, Schwarz, Badeer, Swee, Halpern, Herman, Sizov, Guangyi, Zhang, Lakshminarayanan, Inan, Shojanazeri, Zou, Wang, Zha, Habeeb, Rudolph, Suk, Aspegren, Goldman, Zhan, Damlaj, Molybog, Tufanov, Leontiadis, Veliche, Gat, Weissman, Geboski, Kohli, Lam, Asher, Gaya, Marcus, Tang, Chan, Zhen, Reizenstein, Teboul, Zhong, Jin, Yang, Cummings, Carvill, Shepard, McPhie,
  Torres, Ginsburg, Wang, Wu, U, Saxena, Khandelwal, Zand, Matosich, Veeraraghavan, Michelena, Li, Jagadeesh, Huang, Chawla, Huang, Chen, Garg, A, Silva, Bell, Zhang, Guo, Yu, Moshkovich, Wehrstedt, Khabsa, Avalani, Bhatt, Mankus, Hasson, Lennie, Reso, Groshev, Naumov, Lathi, Keneally, Liu, Seltzer, Valko, Restrepo, Patel, Vyatskov, Samvelyan, Clark, Macey, Wang, Hermoso, Metanat, Rastegari, Bansal, Santhanam, Parks, White, Bawa, Singhal, Egebo, Usunier, Mehta, Laptev, Dong, Cheng, Chernoguz, Hart, Salpekar, Kalinli, Kent, Parekh, Saab, Balaji, Rittner, Bontrager, Roux, Dollar, Zvyagina, Ratanchandani, Yuvraj, Liang, Alao, Rodriguez, Ayub, Murthy, Nayani, Mitra, Parthasarathy, Li, Hogan, Battey, Wang, Howes, Rinott, Mehta, Siby, Bondu, Datta, Chugh, Hunt, Dhillon, Sidorov, Pan, Mahajan, Verma, Yamamoto, Ramaswamy, Lindsay, Lindsay, Feng, Lin, Zha, Patil, Shankar, Zhang, Zhang, Wang, Agarwal, Sajuyigbe, Chintala, Max, Chen, Kehoe, Satterfield, Govindaprasad, Gupta, Deng, Cho, Virk, Subramanian, Choudhury,
  Goldman, Remez, Glaser, Best, Koehler, Robinson, Li, Zhang, Matthews, Chou, Shaked, Vontimitta, Ajayi, Montanez, Mohan, Kumar, Mangla, Ionescu, Poenaru, Mihailescu, Ivanov, Li, Wang, Jiang, Bouaziz, Constable, Tang, Wu, Wang, Wu, Gao, Kleinman, Chen, Hu, Jia, Qi, Li, Zhang, Zhang, Adi, Nam, Yu, Wang, Zhao, Hao, Qian, Li, He, Rait, DeVito, Rosnbrick, Wen, Yang, Zhao, and Ma}]{grattafiori2024llama3herdmodels}
Aaron Grattafiori, Abhimanyu Dubey, Abhinav Jauhri, Abhinav Pandey, Abhishek Kadian, Ahmad Al-Dahle, Aiesha Letman, Akhil Mathur, Alan Schelten, Alex Vaughan, Amy Yang, Angela Fan, Anirudh Goyal, Anthony Hartshorn, Aobo Yang, Archi Mitra, Archie Sravankumar, Artem Korenev, Arthur Hinsvark, and 542 others. 2024.
\newblock \href {https://arxiv.org/abs/2407.21783} {The llama 3 herd of models}.
\newblock \emph{Preprint}, arXiv:2407.21783.

\bibitem[{Guilford(1967)}]{guilford1967nature}
Joy~Paul Guilford. 1967.
\newblock \emph{The Nature of Human Intelligence}.
\newblock McGraw-Hill.

\bibitem[{Guilford et~al.(1978)Guilford, Christensen, Merrifield, and Wilson}]{guildford1978alternate}
JP~Guilford, PR~Christensen, PR~Merrifield, and RC~Wilson. 1978.
\newblock \emph{Alternate Uses: Manual of Instructions and Interpretation}.
\newblock Orange, CA: Sheridan Psychological Services.

\bibitem[{Hofelich~Mohr et~al.(2016)Hofelich~Mohr, Sell, and Lindsay}]{hofelich2016thinking}
Alicia Hofelich~Mohr, Andrew Sell, and Thomas Lindsay. 2016.
\newblock Thinking inside the box: Visual design of the response box affects creative divergent thinking in an online survey.
\newblock \emph{Social Science Computer Review}, 34(3):347--359.

\bibitem[{Izacard and Grave(2020)}]{izacard2020distilling}
Gautier Izacard and Edouard Grave. 2020.
\newblock Distilling knowledge from reader to retriever for question answering.
\newblock \emph{arXiv preprint arXiv:2012.04584}.

\bibitem[{Karwowski(2014)}]{karwowski2014creative}
Maciej Karwowski. 2014.
\newblock Creative mindsets: Measurement, correlates, consequences.
\newblock \emph{Psychology of Aesthetics, Creativity, and the Arts}, 8(1):62.

\bibitem[{Kelty et~al.(2025)Kelty, Baten, Proma, Hoque, Bollen, and Ghoshal}]{kelty2023don}
Sean Kelty, Raiyan~Abdul Baten, Adiba~Mahbub Proma, Ehsan Hoque, Johan Bollen, and Gourab Ghoshal. 2025.
\newblock \href {https://doi.org/10.1057/s41599-025-05124-z} {The innovation trade-off: how following superstars shapes academic novelty}.
\newblock \emph{Humanities and Social Sciences Communications}, 12(1):926.

\bibitem[{Khandelwal et~al.(2020)Khandelwal, Fan, Jurafsky, Zettlemoyer, and Lewis}]{khandelwal2020generalization}
Urvashi Khandelwal, Angela Fan, Dan Jurafsky, Luke Zettlemoyer, and Mike Lewis. 2020.
\newblock Generalization through memorization: Nearest neighbor language models.
\newblock In \emph{Proceedings of the International Conference on Learning Representations (ICLR)}.

\bibitem[{Lee and Ashton(2004)}]{lee2004psychometric}
Kibeom Lee and Michael~C Ashton. 2004.
\newblock Psychometric properties of the {HEXACO} personality inventory.
\newblock \emph{Multivariate Behavioral Research}, 39(2):329--358.

\bibitem[{Lewis et~al.(2020)Lewis, Perez, Piktus, Petroni, Karpukhin, Goyal, Küttler, Lewis, tau Yih, Rocktäschel, Riedel, and Kiela}]{lewis2021retrieval}
Patrick Lewis, Ethan Perez, Aleksandra Piktus, Fabio Petroni, Vladimir Karpukhin, Naman Goyal, Heinrich Küttler, Mike Lewis, Wen tau Yih, Tim Rocktäschel, Sebastian Riedel, and Douwe Kiela. 2020.
\newblock Retrieval-augmented generation for knowledge-intensive {NLP} task.
\newblock \emph{Advances in Neural Information Processing Systems}, 33:9459--9474.

\bibitem[{Li et~al.(2024{\natexlab{a}})Li, Jiang, Huang, Beigi, Zhao, Tan, Bhattacharjee, Jiang, Chen, Wu et~al.}]{li2024generation}
Dawei Li, Bohan Jiang, Liangjie Huang, Alimohammad Beigi, Chengshuai Zhao, Zhen Tan, Amrita Bhattacharjee, Yuxuan Jiang, Canyu Chen, Tianhao Wu, and 1 others. 2024{\natexlab{a}}.
\newblock From generation to judgment: Opportunities and challenges of {LLM}-as-a-judge.
\newblock \emph{arXiv preprint arXiv:2411.16594}.

\bibitem[{Li et~al.(2024{\natexlab{b}})Li, Yang, Tan, Baik, Yun, Lee, Chacko, Hou, Duong-Tran, Ding et~al.}]{li2024dalk}
Dawei Li, Shu Yang, Zhen Tan, Jae~Young Baik, Sukwon Yun, Joseph Lee, Aaron Chacko, Bojian Hou, Duy Duong-Tran, Ying Ding, and 1 others. 2024{\natexlab{b}}.
\newblock {DALK}: Dynamic co-augmentation of {LLM}s and {KG} to answer {A}lzheimer's disease questions with scientific literature.
\newblock \emph{arXiv preprint arXiv:2405.04819}.

\bibitem[{Liang et~al.(2023)Liang, He, Jiao, Wang, Wang, Wang, Yang, Shi, and Tu}]{liang2023encouraging}
Tian Liang, Zhiwei He, Wenxiang Jiao, Xing Wang, Yan Wang, Rui Wang, Yujiu Yang, Shuming Shi, and Zhaopeng Tu. 2023.
\newblock Encouraging divergent thinking in large language models through multi-agent debate.
\newblock \emph{arXiv preprint arXiv:2305.19118}.

\bibitem[{Liu et~al.(2023)Liu, Lin, Hewitt, Paranjape, Bevilacqua, Petroni, and Liang}]{liu2023lost}
Nelson~F. Liu, Kevin Lin, John Hewitt, Ashwin Paranjape, Michele Bevilacqua, Fabio Petroni, and Percy Liang. 2023.
\newblock \href {https://arxiv.org/abs/2307.03172} {Lost in the middle: How language models use long contexts}.
\newblock \emph{Preprint}, arXiv:2307.03172.

\bibitem[{Lu et~al.(2024)Lu, Sclar, Hallinan, Mireshghallah, Liu, Han, Ettinger, Jiang, Chandu, Dziri et~al.}]{lu2024ai}
Ximing Lu, Melanie Sclar, Skyler Hallinan, Niloofar Mireshghallah, Jiacheng Liu, Seungju Han, Allyson Ettinger, Liwei Jiang, Khyathi Chandu, Nouha Dziri, and 1 others. 2024.
\newblock {AI} as humanity's {Salieri}: Quantifying linguistic creativity of language models via systematic attribution of machine text against web text.
\newblock \emph{arXiv preprint arXiv:2410.04265}.

\bibitem[{MacQueen(1967)}]{macqueen1967kmeans}
J.~MacQueen. 1967.
\newblock Some methods for classification and analysis of multivariate observations.
\newblock In \emph{Proceedings of the Fifth Berkeley Symposium on Mathematical Statistics and Probability}, volume~1, pages 281--297. University of California Press.

\bibitem[{McCrae et~al.(2005)McCrae, Costa, and Martin}]{mccrae2005neo}
Robert~R McCrae, Paul~T Costa, Jr, and Thomas~A Martin. 2005.
\newblock The {NEO}-{PI}-3: A more readable revised {NEO} personality inventory.
\newblock \emph{Journal of {P}ersonality {A}ssessment}, 84(3):261--270.

\bibitem[{{Meta AI}(2024)}]{llama3huggingface}
{Meta AI}. 2024.
\newblock {LLaMA 3.3-70B-Instruct}.
\newblock \url{https://huggingface.co/meta-llama/Llama-3.3-70B-Instruct}.
\newblock Accessed: 2025-05-18.

\bibitem[{Mimno et~al.(2011)Mimno, Wallach, Talley, Leenders, and McCallum}]{mimno2011optimizing}
David Mimno, Hanna Wallach, Edmund Talley, Miriam Leenders, and Andrew McCallum. 2011.
\newblock Optimizing semantic coherence in topic models.
\newblock In \emph{Proceedings of the 2011 Conference on Empirical Methods in Natural Language Processing}, pages 262--272.

\bibitem[{Newman(2018)}]{newman2018networks}
Mark Newman. 2018.
\newblock \emph{Networks}.
\newblock Oxford University Press.

\bibitem[{Olson et~al.(2021)Olson, Nahas, Chmoulevitch, Cropper, and Webb}]{olson2021naming}
Jay~A Olson, Johnny Nahas, Denis Chmoulevitch, Simon~J Cropper, and Margaret~E Webb. 2021.
\newblock Naming unrelated words predicts creativity.
\newblock \emph{Proceedings of the National Academy of Sciences}, 118(25):e2022340118.

\bibitem[{Organisciak et~al.(2023)Organisciak, Acar, Dumas, and Berthiaume}]{organisciak2023beyond}
Peter Organisciak, Selcuk Acar, Denis Dumas, and Kelly Berthiaume. 2023.
\newblock Beyond semantic distance: Automated scoring of divergent thinking greatly improves with large language models.
\newblock \emph{Thinking Skills and Creativity}, 49:101356.

\bibitem[{Organisciak and Dumas(2020)}]{organisciak2020open}
Peter Organisciak and Denis Dumas. 2020.
\newblock Open creativity scoring.
\newblock \url{https://openscoring.du.edu}.
\newblock [Computer software].

\bibitem[{Page(2018)}]{page2018model}
Scott~E. Page. 2018.
\newblock \emph{The Model Thinker: What You Need to Know to Make Data Work for You}.
\newblock Basic Books, New York.

\bibitem[{Papineni et~al.(2002)Papineni, Roukos, Ward, and Zhu}]{papineni2002bleu}
Kishore Papineni, Salim Roukos, Todd Ward, and Wei-Jing Zhu. 2002.
\newblock {BLEU}: a method for automatic evaluation of machine translation.
\newblock In \emph{Proceedings of the 40th Annual Meeting of the Association for Computational Linguistics}, pages 311--318.

\bibitem[{Reimers and Gurevych(2019)}]{reimers2019sentencebert}
Nils Reimers and Iryna Gurevych. 2019.
\newblock \href {https://arxiv.org/abs/1908.10084} {Sentence-{BERT}: Sentence embeddings using siamese {BERT}-networks}.
\newblock In \emph{Proceedings of the 2019 Conference on Empirical Methods in Natural Language Processing}. Association for Computational Linguistics.

\bibitem[{Reiter-Palmon et~al.(2019)Reiter-Palmon, Forthmann, and Barbot}]{reiter2019scoring}
Roni Reiter-Palmon, Boris Forthmann, and Baptiste Barbot. 2019.
\newblock Scoring divergent thinking tests: A review and systematic framework.
\newblock \emph{Psychology of Aesthetics, Creativity, and the Arts}, 13(2):144.

\bibitem[{Rosenberg and Hirschberg(2007)}]{rosenberg2007v}
Andrew Rosenberg and Julia Hirschberg. 2007.
\newblock \href {https://aclanthology.org/D07-1043/} {{V}-measure: A conditional entropy-based external cluster evaluation measure}.
\newblock In \emph{Proceedings of the 2007 Joint Conference on Empirical Methods in Natural Language Processing and Computational Natural Language Learning ({EMNLP}-{C}o{NLL})}, pages 410--420, Prague, Czech Republic. Association for Computational Linguistics.

\bibitem[{Rousseeuw(1987)}]{rousseeuw1987silhouettes}
Peter~J Rousseeuw. 1987.
\newblock Silhouettes: A graphical aid to the interpretation and validation of cluster analysis.
\newblock \emph{Journal of Computational and Applied Mathematics}, 20:53--65.

\bibitem[{Runco and Jaeger(2012)}]{runco2012standard}
Mark~A Runco and Garrett~J Jaeger. 2012.
\newblock The standard definition of creativity.
\newblock \emph{Creativity Research Journal}, 24(1):92--96.

\bibitem[{Runco and Mraz(1992)}]{runco1992scoring}
Mark~A Runco and Wayne Mraz. 1992.
\newblock Scoring divergent thinking tests using total ideational output and a creativity index.
\newblock \emph{Educational and Psychological Measurement}, 52(1):213--221.

\bibitem[{Shen et~al.(2024)Shen, Wang, Shi, Du, Tao, Shen, and Zhang}]{shen2024comparative}
Yanxin Shen, Lun Wang, Chuanqi Shi, Shaoshuai Du, Yiyi Tao, Yixian Shen, and Hang Zhang. 2024.
\newblock Comparative analysis of listwise reranking with large language models in limited-resource language contexts.
\newblock \emph{arXiv preprint arXiv:2412.20061}.

\bibitem[{Shrout and Fleiss(1979)}]{shrout1979intraclass}
Patrick~E Shrout and Joseph~L Fleiss. 1979.
\newblock Intraclass correlations: Uses in assessing rater reliability.
\newblock \emph{Psychological Bulletin}, 86(2):420.

\bibitem[{Silvia et~al.(2017)Silvia, Nusbaum, and Beaty}]{silvia2017old}
Paul~J Silvia, Emily~C Nusbaum, and Roger~E Beaty. 2017.
\newblock Old or new? {E}valuating the old/new scoring method for divergent thinking tasks.
\newblock \emph{The Journal of Creative Behavior}, 51(3):216--224.

\bibitem[{Silvia et~al.(2008)Silvia, Winterstein, Willse, Barona, Cram, Hess, Martinez, and Richard}]{silvia2008assessing}
Paul~J Silvia, Beate~P Winterstein, John~T Willse, Christopher~M Barona, Joshua~T Cram, Karl~I Hess, Jenna~L Martinez, and Crystal~A Richard. 2008.
\newblock Assessing creativity with divergent thinking tasks: Exploring the reliability and validity of new subjective scoring methods.
\newblock \emph{Psychology of Aesthetics, Creativity, and the Arts}, 2(2):68.

\bibitem[{Snyder et~al.(2004)Snyder, Mitchell, Bossomaier, and Pallier}]{snyder2004creativity}
Allan Snyder, John Mitchell, Terry Bossomaier, and Gerry Pallier. 2004.
\newblock The {C}reativity {Q}uotient: An objective scoring of ideational fluency.
\newblock \emph{Creativity Research Journal}, 16(4):415--419.

\bibitem[{Stevenson et~al.(2020)Stevenson, Smal, Baas, Dahrendorf, Grasman, Tanis, Scheurs, Sleiffer, and van~der Maas}]{stevenson2020automated}
C.~Stevenson, I.~Smal, M.~Baas, M.~Dahrendorf, R.~Grasman, C.~Tanis, E.~Scheurs, D.~Sleiffer, and H.~van~der Maas. 2020.
\newblock \href {https://hdl.handle.net/11245.1/13ad004a-1b61-45a0-8a9a-56d7a165d7ef} {Automated {AUT} scoring using a big data variant of the consensual assessment technique}.
\newblock Report Final Technical Report, Modeling Creativity Project, Universiteit van Amsterdam, Amsterdam.
\newblock Faculty of Social and Behavioural Sciences (FMG), Psychology Research Institute (PsyRes).

\bibitem[{Tanveer et~al.(2018)Tanveer, Samrose, Baten, and Hoque}]{tanveer2018awe}
M~Iftekhar Tanveer, Samiha Samrose, Raiyan~Abdul Baten, and M~Ehsan Hoque. 2018.
\newblock Awe the audience: How the narrative trajectories affect audience perception in public speaking.
\newblock In \emph{Proceedings of the 2018 CHI Conference on Human Factors in Computing Systems}, pages 1--12.

\bibitem[{Thurstone(1938)}]{Ces_1938}
L.~L. Thurstone. 1938.
\newblock \href {https://doi.org/10.2307/3607923} {Primary mental abilities}.
\newblock \emph{The Mathematical Gazette}, 22(251):411–412.

\bibitem[{Vinh et~al.(2010)Vinh, Epps, and Bailey}]{Vinh2010}
Nguyen~Xuan Vinh, Julien Epps, and James Bailey. 2010.
\newblock \href {http://jmlr.org/papers/v11/vinh10a.html} {Information theoretic measures for clusterings comparison: Variants, properties, normalization and correction for chance}.
\newblock \emph{Journal of Machine Learning Research}, 11(95):2837--2854.

\bibitem[{Viswanathan et~al.(2024)Viswanathan, Gashteovski, Gashteovski, Lawrence, Wu, and Neubig}]{Viswanathan2024}
Vijay Viswanathan, Kiril Gashteovski, Kiril Gashteovski, Carolin Lawrence, Tongshuang Wu, and Graham Neubig. 2024.
\newblock \href {https://doi.org/10.1162/tacl_a_00648} {Large language models enable few-shot clustering}.
\newblock \emph{Transactions of the Association for Computational Linguistics}, 12:321--333.

\bibitem[{Wang et~al.(2022)Wang, Yang, Huang, Jiao, Yang, Jiang, Majumder, and Wei}]{wang2022text}
Liang Wang, Nan Yang, Xiaolong Huang, Binxing Jiao, Linjun Yang, Daxin Jiang, Rangan Majumder, and Furu Wei. 2022.
\newblock Text embeddings by weakly-supervised contrastive pre-training.
\newblock \emph{arXiv preprint arXiv:2212.03533}.

\bibitem[{Ward~Jr(1963)}]{ward1963hierarchical}
Joe~H. Ward~Jr. 1963.
\newblock Hierarchical grouping to optimize an objective function.
\newblock \emph{Journal of the American Statistical Association}, 58(301):236--244.

\bibitem[{Wei et~al.(2022)Wei, Wang, Schuurmans, Bosma, Ichter, Xia, Chi, Le, and Zhou}]{wei2022chain}
Jason Wei, Xuezhi Wang, Dale Schuurmans, Maarten Bosma, Brian Ichter, Fei Xia, Ed~Chi, Quoc Le, and Denny Zhou. 2022.
\newblock Chain of thought prompting elicits reasoning in large language models.
\newblock \emph{Advances in Neural Information Processing Systems (NeurIPS)}.

\bibitem[{Xiao et~al.(2023{\natexlab{a}})Xiao, Liu, Zhang, and Muennighoff}]{bge_embedding}
Shitao Xiao, Zheng Liu, Peitian Zhang, and Niklas Muennighoff. 2023{\natexlab{a}}.
\newblock \href {https://arxiv.org/abs/2309.07597} {C-pack: Packaged resources to advance general chinese embedding}.
\newblock \emph{Preprint}, arXiv:2309.07597.

\bibitem[{Xiao et~al.(2023{\natexlab{b}})Xiao, Yuan, Liao, Abdelghani, and Oudeyer}]{xiao2023supporting}
Ziang Xiao, Xingdi Yuan, Q~Vera Liao, Rania Abdelghani, and Pierre-Yves Oudeyer. 2023{\natexlab{b}}.
\newblock Supporting qualitative analysis with large language models: Combining codebook with {GPT}-3 for deductive coding.
\newblock In \emph{Companion Proceedings of the 28th International Conference on Intelligent User Interfaces}, pages 75--78.

\bibitem[{Yang et~al.(2025)Yang, Li, Yang, Zhang, Hui, Zheng, Yu, Gao, Huang, Lv, Zheng, Liu, Zhou, Huang, Hu, Ge, Wei, Lin, Tang, Yang, Tu, Zhang, Yang, Yang, Zhou, Zhou, Lin, Dang, Bao, Yang, Yu, Deng, Li, Xue, Li, Zhang, Wang, Zhu, Men, Gao, Liu, Luo, Li, Tang, Yin, Ren, Wang, Zhang, Ren, Fan, Su, Zhang, Zhang, Wan, Liu, Wang, Cui, Zhang, Zhou, and Qiu}]{yang2025qwen3technicalreport}
An~Yang, Anfeng Li, Baosong Yang, Beichen Zhang, Binyuan Hui, Bo~Zheng, Bowen Yu, Chang Gao, Chengen Huang, Chenxu Lv, Chujie Zheng, Dayiheng Liu, Fan Zhou, Fei Huang, Feng Hu, Hao Ge, Haoran Wei, Huan Lin, Jialong Tang, and 41 others. 2025.
\newblock \href {https://arxiv.org/abs/2505.09388} {Qwen3 technical report}.
\newblock \emph{Preprint}, arXiv:2505.09388.

\bibitem[{Yao et~al.(2023)Yao, Yu, Zhao, Shafran, Griffiths, Cao, and Narasimhan}]{yao2023tree}
Shunyu Yao, Dian Yu, Jeffrey Zhao, Izhak Shafran, Tom Griffiths, Yuan Cao, and Karthik Narasimhan. 2023.
\newblock Tree of thoughts: Deliberate problem solving with large language models.
\newblock \emph{Advances in Neural Information Processing Systems}, 36:11809--11822.

\bibitem[{Zhang et~al.(2019)Zhang, Kishore, Wu, Weinberger, and Artzi}]{zhang2019bertscore}
Tianyi Zhang, Varsha Kishore, Felix Wu, Kilian~Q Weinberger, and Yoav Artzi. 2019.
\newblock {BERT}score: Evaluating text generation with {BERT}.
\newblock \emph{arXiv preprint arXiv:1904.09675}.

\bibitem[{Zhao et~al.(2024)Zhao, Yang, Zhang, Shao, Zhang, Qiao, Luo, and Ji}]{zhao2024diffagent}
Lirui Zhao, Yue Yang, Kaipeng Zhang, Wenqi Shao, Yuxin Zhang, Yu~Qiao, Ping Luo, and Rongrong Ji. 2024.
\newblock Diffagent: Fast and accurate text-to-image {API} selection with large language model.
\newblock In \emph{Proceedings of the IEEE/CVF Conference on Computer Vision and Pattern Recognition}, pages 6390--6399.

\end{thebibliography}
\end{document}